\documentclass{article}

\usepackage{arxiv}

\usepackage[utf8]{inputenc} 
\usepackage[T1]{fontenc}    
\usepackage{hyperref}       
\usepackage{url}            
\usepackage{booktabs}       
\usepackage{amsfonts}       
\usepackage{nicefrac}       
\usepackage{microtype}      
\usepackage{lipsum}

\usepackage{amsmath}
\usepackage{amssymb}
\usepackage{bbm}
\usepackage{mathtools}
\DeclareMathOperator{\SoftMax}{SoftMax}

\usepackage{url}

\usepackage[edges]{forest}
\usepackage{tikz}
\usetikzlibrary{arrows.meta,shadows.blur}
\usetikzlibrary{shapes.arrows}

\usepackage{tabularx}
\usepackage{float} 
\usepackage{booktabs,multirow,threeparttable}
\usepackage{subcaption}

\title{A Survey of Techniques All Classifiers Can Learn from Deep Networks: Models, Optimizations, and Regularization}

\author{
  Alireza Ghods \\
  Department of Computer Science\\
  Washington State University\\
  Pullman, WA 99163 \\
  \texttt{alireza.ghods@wsu.edu} \\
   \And
 Diane J.~Cook \\
  Department of Computer Science\\
  Washington State University\\
  Pullman, WA 99163 \\
  \texttt{djcook@wsu.edu} \\
}

\begin{document}
\maketitle

\begin{abstract}
Deep neural networks have introduced novel and useful tools to the machine learning community. Other types of classifiers can potentially make use of these tools as well to improve their performance and generality. This paper reviews the current state of the art for deep learning classifier technologies that are being used outside of deep neural networks. Non-network classifiers can employ many components found in deep neural network architectures. In this paper, we review the feature learning, optimization, and regularization methods that form a core of deep network technologies. We then survey non-neural network learning algorithms that make innovative use of these methods to improve classification. Because many opportunities and challenges still exist, we discuss directions that can be pursued to expand the area of deep learning for a variety of classification algorithms.
\end{abstract}



\section{Introduction}

The objective of supervised learning algorithms is to identify an optimal mapping between input features and output values based on a given training dataset. A supervised learning method that is attracting substantial research and industry attention is Deep Neural Networks (DNN). DNNs have a profound effect on our daily lives; they are found in search engines \cite{wired2016} \cite{tc2015}, self-driving cars \cite{wired2017} \cite{nytimes2017} \cite{cadmv}, health care systems \cite{jiang2017artificial}, and consumer devices such as smart-phones and cameras \cite{google2018}. Convolutional Neural Networks (CNN) have become the standard for processing images \cite{wan2013regularization} \cite{graham2014fractional} \cite{clevert2015fast}, whereas Recurrent Neural Networks (RNN) dominate the processing of sequential data such as text and voice \cite{hermann2015teaching} \cite{lample2016neural} \cite{bahdanau2016end} \cite{amodei2016deep}. DNNs allow machines to automatically discover the representations needed for detection or classification of raw input \cite{lecun2015deep}. Additionally, the neural network community developed unsupervised algorithms to help with the learning of unlabeled data. These unsupervised methods have found their way to real-world applications, such as creating generative adversarial networks (GANs) that design clothes \cite{VERGE2017}. The term \textbf{\textit{deep}} has been used to distinguish these networks from shallow networks which have only one hidden layer; in contrast, DNNs have multiple hidden layers. The two terms \textbf{\textit{deep learning}} and \textbf{\textit{deep neural networks}} have been used synonymously. However, we observe that deep leaning itself conveys a broader meaning, which can also shape the field of machine learning outside the realm of neural network algorithms.

The remarkable recent DNN advances were made possible by the availability of massive amounts of computational power and labeled data. However, these advances do not overcome all of the difficulties associated with DNNs. For example, there are many real-world scenarios, such as analyzing power distribution data \cite{tang2018efficient}, for which large annotated datasets do not exist due to the complexity and expense of collecting data. While applications like clinical interpretation of medical diagnoses require that the learned model be understandable, most DNNs resist interpretation due to their complexity  \cite{caruana2015intelligible}. DNNs can be insensitive to noisy training data \cite{nguyen2015deep} \cite{zhang2016understanding} \cite{krueger2017deep}, and they also require appropriate parameter initialization to converge \cite{mishkin2015all} \cite{kumar2017weight}.

Despite these shortcomings, DNNs have reported higher predictive accuracy than other supervised learning methods for many datasets, given enough supervised data and computational resources. Deep models offer structural advantages that may improve the quality of learning in complex datasets as empirically shown by Bengio \cite{bengio2009learning}. Recently, researchers have designed hybrid methods which combine unique DNN techniques with other classifiers to address some of these identified problems or to boost other classifiers. This survey paper investigates these methods, reviewing classifiers which have adapted DNN techniques to alternative classifiers.

\subsection{Research Objectives and Outline}

While DNN research is growing rapidly, this paper aims to draw a broader picture of deep learning methods. Although some studies provide evidence that DNN models offer greater generalization than classic machine learning algorithms for complex data  \cite{szegedy2015going} \cite{wu2016google} \cite{jozefowicz2016exploring} \cite{graves2013speech} \cite{ji20133d}, there is no ``silver bullet'' approach to concept learning \cite{wolpert1997no}. Numerous studies comparing DNNs and other supervised learning algorithms \cite{king1995statlog} \cite{lim2000comparison} \cite{caruana2006empirical} \cite{caruana2008empirical} \cite{baumann2019comparative} observe that the choice of algorithm depends on the data - no ideal algorithm exists which generalizes optimally on all types of data. Recognizing the unique and important role other classifiers thus play, we aim to investigate how non-network machine learning algorithms can benefit from the advances in deep neural networks. Many deep learning survey papers have been published that provide a primer on the topic \cite{pouyanfar2018survey} or highlight diverse applications such as object detection \cite{shickel2018deep}, medical record analysis \cite{han2018advanced}, activity recognition \cite{wang2019deep}, and natural language processing  \cite{hatcher2018survey}. In this survey, we do not focus solely on deep neural network models but rather on how deep learning can inspire a broader range of classifiers. We concentrate on research breakthroughs that transform non-network classifiers into deep learners. Further, we review deep network techniques such as stochastic gradient descent that can be used more broadly, and we discuss ways in which non-network models can benefit from network-inspired deep learning innovations. 

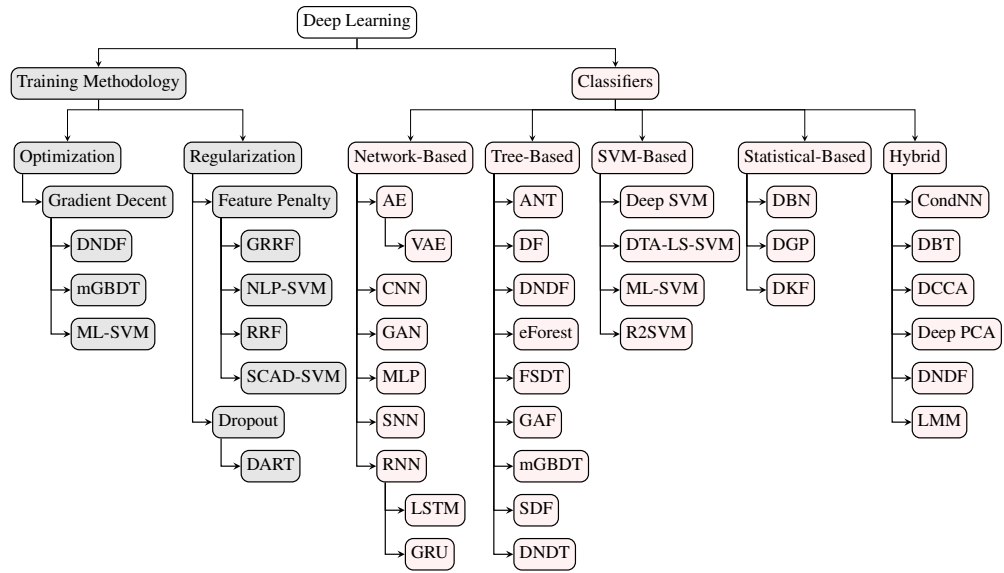
\begin{figure*}[h]
    \centering
    \resizebox{\textwidth}{!}{
    \begin{forest}
        for tree={
            font=\footnotesize,
            draw,
            semithick,
            rounded corners,
            align = center,
            inner sep = 1mm,
            edge = {semithick, -stealth},
            l sep=0.5cm,
            s sep=0.2cm,
            fork sep=1mm,
            parent anchor=south,
            child anchor=north,
            edge path={
              \noexpand\path [-{Stealth[]}, \forestoption{edge}, thin]
                (!u.parent anchor) -- +(0,-5pt) -| (.child anchor)\forestoption{edge label};
            },
            /tikz/>=LaTeX,
        },
        [Deep Learning, 
            [Training Methodology, for tree={fill=gray!20}
                [Optimization, for tree={folder,grow'=0}
                    [Gradient Decent 
                        [DNDF]
                        [mGBDT]
                        [ML-SVM]
                    ]
                ]
                [Regularization, for tree={folder,grow'=0}
                    [Feature Penalty
                        [GRRF]
                        [NLP-SVM]
                        [RRF]
                        [SCAD-SVM]
                    ]
                    [Dropout
                        [DART]
                    ]
                ]
            ]
            [Classifiers, for tree={fill=pink!20}
                [Network-Based, for tree={folder,grow'=0}
                    [AE
                        [VAE]
                    ]
                    [CNN]
                    [GAN]
                    [MLP]
                    [SNN]
                    [RNN
                        [LSTM]
                        [GRU]
                    ]
                ]
                [Tree-Based, for tree={folder,grow'=0}
                    [ANT]
                    [DF]
                    [DNDF]
                    [eForest]
                    [FSDT]
                    [GAF]
                    [mGBDT]
                    [SDF]
                    [DNDT]
                ]
                [SVM-Based, for tree={folder,grow'=0}
                    [Deep SVM]
                    [DTA-LS-SVM]
                    [ML-SVM]
                    [R2SVM]
                ]
                [Statistical-Based, for tree={folder,grow'=0}
                    [DBN]
                    [DGP]
                    [DKF]
                ]
                [Hybrid, for tree={folder,grow'=0}
                    [CondNN]
                    [DBT]
                    [DCCA]
                    [Deep PCA]
                    [DNDF]
                    [LMM]
                ]
            ] 
        ]
    \end{forest}}
    \caption{Content map of the methods covered in this survey.}
    \label{fig:overview}
\end{figure*}

The literature provides evidence that non-network models may offer improved generalizability over deep networks, depending on the amount and type of data that is available. By surveying methods for transforming non-network classifiers into deep learners, these approaches can become stronger learners. To provide evidence of the need for continued research on this topic, we also implement a collection of shallow and deep learners surveyed in this paper, both network and non-network classifiers, to compare their performance. Figure \ref{fig:overview} highlights deep learning components that we discuss in this survey. This graph also summarizes the deep classifiers that we survey and the relationships that we highlight between techniques.

\section{Brief Overview of Deep Neural Networks}

\subsection{The Origin}

In 1985, Rosenblatt introduced the Perceptron \cite{rosenblatt1958perceptron}, an online binary classifier which flows input through a weight vector to an output layer. Perceptron learning uses a form of gradient descent to adjust the weights between the input and output layers to optimize a loss function \cite{widrow1960adaptive}. A few years later, Minsky proved that a single-layer Perceptron is unable to learn nonlinear functions, including the XOR function \cite{minsky2017perceptrons}. Multilayer perceptrons (MLPs, see Table~\ref{tab:abb_dis} for a complete list of abbreviations) addressed the nonlinearity problem by adding layers of hidden units to the networks and applying alternative differentiable activation functions, such as sigmoid, to each node. Stochastic gradient descent was then applied to MLPs to determine the weights between layers that minimize function approximation errors \cite{rumelhart1985learning}. However, the lack of computational power caused DNN research to stagnate for decades, and other classifiers rose in popularity. In 2006, a renaissance began in DNN research, spurred by the introduction of Deep Belief Networks (DBNs) \cite{hinton2006fast}. 

\subsection{Deep Neural Network Architectures}

Due to the increasing popularity of deep learning, many DNN architectures have been introduced with variations such as Neural Turing Machines \cite{graves2014neural} and Capsule Neural Networks \cite{NIPS2017_6975}. In this paper, we summarize the general form of DNNs together with architectural components that not only appear in DNNs but can be incorporated into other models. We start by reviewing popular types of DNNs that have been introduced and that play complementary learning roles.

\subsection{Supervised Learning}

\subsubsection{Multilayer Perceptron}

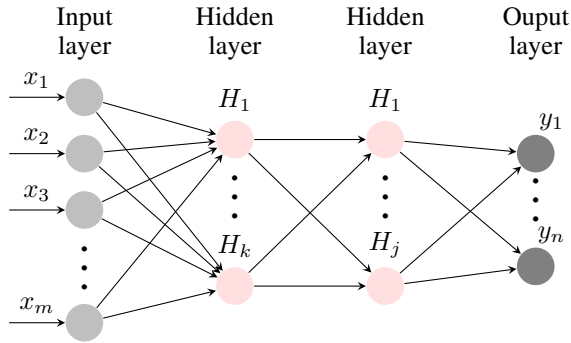
\begin{figure}[h]
   \centering
    \tikzset{%
       neuron missing/.style={
        draw=none, 
        scale=1.8,
        text height=0.333cm,
        execute at begin node=\color{black}$\vdots$
      },
    }
    
    \begin{tikzpicture}[x=1cm, y=.75cm, >=stealth]
    
        \foreach \m/\l [count=\y] in {1,2,3} {
         \node [circle,fill=gray!50,minimum size=.5cm] (input-\m) at (0,2.5-\y) {};
        }
        \foreach \m/\l [count=\y] in {4} {
         \node [circle,fill=gray!50,minimum size=.5cm ] (input-\m) at (0,-2.5) {};
        }
         
        \node [neuron missing]  at (0,-1.5) {};
         
        \foreach \m [count=\y] in {1}
          \node [circle,fill=pink!50,minimum size=.5cm ] (hidden1-\m) at (2,0.75) {};
          
        \foreach \m [count=\y] in {2}
          \node [circle,fill=pink!50,minimum size=.5cm ] (hidden1-\m) at (2,-1.85) {};
          
         \node [neuron missing]  at (2,-0.25) {};
         
         \foreach \m [count=\y] in {1}
          \node [circle,fill=pink!50,minimum size=.5cm ] (hidden2-\m) at (4,0.75) {};
          
        \foreach \m [count=\y] in {2}
          \node [circle,fill=pink!50,minimum size=.5cm ] (hidden2-\m) at (4,-1.85) {};
          
         \node [neuron missing]  at (4,-0.25) {};
        
        \foreach \m [count=\y] in {1}
          \node [circle,fill=black!50,minimum size=.5cm ] (output-\m) at (6,1.5-\y) {};
          
        \foreach \m [count=\y] in {2}
          \node [circle,fill=black!50,minimum size=0.5cm ] (output-\m) at (6,-0.5-\y) {};
        
         \node [neuron missing]  at (6,-0.3) {};
        
        \foreach \l [count=\i] in {1,2,3,m}
          \draw [<-] (input-\i) -- ++(-1,0)
            node [above, midway] {$x_{\l}$};
        
        \foreach \l [count=\i] in {1,k}
          \node [above] at (hidden1-\i.north) {$H_{\l}$};
          
        \foreach \l [count=\i] in {1,j}
          \node [above] at (hidden2-\i.north) {$H_{\l}$};
          
        \foreach \l [count=\i] in {1,n}
          \node [above, xshift=6pt, yshift=-2pt] at (output-\i.north) {$y_{\l}$};
        
        \foreach \i in {1,...,4}
          \foreach \j in {1,...,2}
            \draw [->] (input-\i) -- (hidden1-\j);
            
        \foreach \i in {1,...,2}
          \foreach \j in {1,...,2}
            \draw [->] (hidden1-\i) -- (hidden2-\j);
        
        \foreach \i in {1,...,2}
          \foreach \j in {1,...,2}
            \draw [->] (hidden2-\i) -- (output-\j);
        
        \foreach \l [count=\x from 0] in {Input, Hidden, Hidden, Ouput}
          \node [align=center, above] at (\x*2,2) {\l \\ layer};
        
    \end{tikzpicture}
    \caption{An illustration of a three-layered MLP with $j$ nodes at the first hidden layer and $k$ at the second layer.}
    \label{fig:MLP}
\end{figure}

A multilayer perceptron (MLP) is one of the essential bases of many deep learning algorithms. The goal of a MLP is to map input $X$ to class $y$ by learning a function $y=f(X, \theta)$, where $\theta$ represents the best possible function approximation. For example, in Figure \ref{fig:MLP} the MLP maps input $X$ to $y$ using function $f(x) = f^{(3)}(f^{(2)}(f^{(1)}(x)))$, where $f^{(1)}$ is the first layer, $f^{(2)}$ is the second layer, and $f^{(3)}$ represents the third, output layer. This chain structure is a common component of many DNN architectures. The network depth is equal to the length of the chain, and the width of each layer represents the number of nodes in that layer \cite{goodfellow2016deep}.

In networks such as the MLP, the connections are not cyclic and thus belong to a class of DNNs called {\em feedforward networks}. Feedforward networks move information in only one direction, from the input to the output layer. Figure~\ref{fig:MLP} depicts a particular type of feedforward network which is a fully-connected multilayer perceptron because each node at one layer is connected to all of the nodes at the next layer. Special cases of feedforward networks and MLPs have drawn considerable recent attention, which we describe next.

\subsubsection{Deep Convolutional Neural Network}

A convolutional neural network (CNN) \cite{lecun1989generalization} is a specialized class of feedforward DNNs for processing data that can be discretely presented. Examples of data that can benefit from CNNs include time series data that can be presented as samples of discrete regular time intervals and image data presented as samples of 2-D pixels at discrete locations. Most CNNs involve three stages: a convolution operation; an activation function, such as the rectified linear activation (ReLU) function \cite{NIPS2012_4824}; and a pooling function, such as max pooling \cite{zhou1988computation}. A convolution operation is a weighted average or smooth estimation of a windowed input. One of the strengths of the convolution operation is that the connections between nodes in a network become sparser by learning a small kernel for unimportant features. Another benefit of convolution is parameter sharing. A CNN makes an assumption that a kernel learned for one input position can be used at every position, in contrast to a MLP which deploys a separate element of a weight matrix for each connection. Applying the convolution operator frequently improves the network's learning ability.

\begin{figure}[h]
	\centering
	\begin{tikzpicture}
		\node at (0.4,-0.5){\begin{tabular}{c}Input\\layer \end{tabular}};
		
		\draw (0,0) -- (0.75,0) -- (0.75,0.75) -- (0,.75) -- (0,0);
		
		\node at (2,-.5){\begin{tabular}{c}Convolutional \\layer + ReLU \end{tabular}};
		
		\draw[fill=black,opacity=0.2,draw=black] (3.25,1.25) -- (3.25,1.25) -- (3.25,2.25) -- (2.25,2.25) -- (2.25,1.25);
		\draw[fill=black,opacity=0.2,draw=black] (3,1) -- (3,1) -- (3,2) -- (2,2) -- (2.,1);
		\draw[fill=black,opacity=0.2,draw=black] (2.75,0.75) -- (2.75,0.75) -- (2.75,1.75) -- (1.75,1.75) -- (1.75,0.75);
		\draw[fill=black,opacity=0.2,draw=black] (2.5,0.5) -- (2.5,0.5) -- (2.5,1.5) -- (1.5,1.5) -- (1.5,0.5);
		\draw[fill=black,opacity=0.2,draw=black] (1.25,0.25) -- (2.25,0.25) -- (2.25,1.25) -- (1.25,1.25) -- (1.25,0.25);
		\draw[fill=black,opacity=0.2,draw=black] (1,0) -- (2,0) -- (2,1) -- (1,1) -- (1,0);
		
		\node at (4.5,2.5){\begin{tabular}{c}Max Pooling\\ layer\end{tabular}};
		
		\draw[fill=black,opacity=0.2,draw=black] (3.75,1.25) -- (5.,1.25) -- (5,2) -- (4.25,2) -- (4.25,1.25);
		\draw[fill=black,opacity=0.2,draw=black] (3.5,1) -- (4.75,1) -- (4.75,1.75) -- (4.,1.75) -- (4.,1);
		\draw[fill=black,opacity=0.2,draw=black] (3.25,0.75) -- (4.5,0.75) -- (4.5,1.5) -- (3.75,1.5) -- (3.75,0.75);
		\draw[fill=black,opacity=0.2,draw=black] (3.5,0.5) -- (4.25,0.5) -- (4.25,1.25) -- (3.5,1.25) -- (3.5,0.5);
		\draw[fill=black,opacity=0.2,draw=black] (3.25,0.25) -- (4.,0.25) -- (4.,1) -- (3.25,1) -- (3.25,0.25);
		\draw[fill=black,opacity=0.2,draw=black] (3,0) -- (3.75,0) -- (3.75,0.75) -- (3,0.75) -- (3,0);
		
		\node at (5.5,-.5){\begin{tabular}{c}Convolutional\\layer + ReLU \end{tabular}};
		
		\draw[fill=black,opacity=0.2,draw=black] (6.25,1.75) -- (6.85,1.75) -- (6.85,2.4) -- (6.25,2.4) -- (6.25,1.75);
		\draw[fill=black,opacity=0.2,draw=black] (6,1.5) -- (6.6,1.5) -- (6.6,2.15) -- (6.,2.15) -- (6.,1.5);
		\draw[fill=black,opacity=0.2,draw=black] (5.75,1.25) -- (6.35,1.25) -- (6.35,1.9) -- (5.75,1.9) -- (5.75,1.25);
		\draw[fill=black,opacity=0.2,draw=black] (5.5,1) -- (6.1,1) -- (6.1,1.65) -- (5.5,1.65) -- (5.5,1);
		\draw[fill=black,opacity=0.2,draw=black] (5.25,0.75) -- (5.85,0.75) -- (5.85,1.4) -- (5.25,1.4) -- (5.25,0.75);
		\draw[fill=black,opacity=0.2,draw=black] (5.,0.5) -- (5.6,0.5) -- (5.6,1.15) -- (5.,1.15) -- (5.,0.5);
		\draw[fill=black,opacity=0.2,draw=black] (4.75,0.25) -- (5.35,0.25) -- (5.35,.9) -- (4.75,.9) -- (4.75,0.25);
		\draw[fill=black,opacity=0.2,draw=black] (4.5,0) -- (5.15,0) -- (5.15,0.65) -- (4.5,0.65) -- (4.5,0);
		
		\node at (7.5,2.5){\begin{tabular}{c}Max Pooling\\layer\end{tabular}};
		
		\draw[fill=black,opacity=0.2,draw=black] (7.25,1.5) -- (7.75,1.5) -- (7.75,2) -- (7.25,2) -- (7.25,1.5);
		\draw[fill=black,opacity=0.2,draw=black] (7.,1.25) -- (7.5,1.25) -- (7.5,1.75) -- (7.,1.75) -- (7.,1.25);
		\draw[fill=black,opacity=0.2,draw=black] (6.75,1) -- (7.25,1) -- (7.25,1.5) -- (6.75,1.5) -- (6.75,1.5);
		\draw[fill=black,opacity=0.2,draw=black] (6.5,0.75) -- (7.,0.75) -- (7.,1.25) -- (6.5,1.25) -- (6.5,0.75);
		\draw[fill=black,opacity=0.2,draw=black] (6.25,0.5) -- (6.75,0.5) -- (6.75,1) -- (6.25,1) -- (6.25,0.5);
		\draw[fill=black,opacity=0.2,draw=black] (6,0.25) -- (6.5,0.25) -- (6.5,0.75) -- (6.,0.75) -- (6.,0.25);
		\draw[fill=black,opacity=0.2,draw=black] (5.75,0) -- (6.25,0) -- (6.25,0.5) -- (5.75,0.5) -- (5.75,0);
		
		\node at (7,-.5){\begin{tabular}{c} FC \\layer \end{tabular}};
		
		\draw[fill=pink,draw=black,opacity=0.5] (6.75,0) -- (7.,0) -- (8.2,1.5) -- (7.95,1.5) -- (6.75,0);
		
	\end{tikzpicture}
	\caption{An illustration of a three-layered CNN made of six convolution filters followed by six max pooling filters at the first layer, and eight convolution filters followed by seven max pooling filters at the second layer. The last layer is a fully connected layer (FC).}
	\label{fig:CNN}
\end{figure}
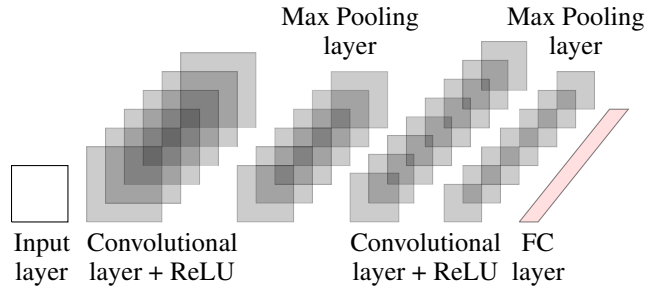

A pooling function replaces the output of specific nearby nodes by their statistical summary. For example, the max-pooling function returns the maximum of a rectangular neighborhood. The motivation behind adding a pooling layer is that statistically down-sampling the number of features makes the representation approximately invariant to small translations of the input by maintaining the essential features. The final output of the learner is generated via a Fully-Connected (FC) layer that appears after the convolutional and max-pooling layers (see Figure~\ref{fig:CNN} for an illustration of the process).

\subsubsection{Recurrent Neural Network} \label{sec:rnn}

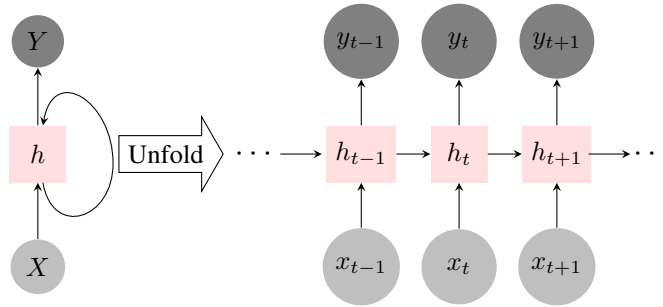
\begin{figure}[h]
    \centering
    \tikzset{%
       neuron missing/.style={
        draw=none, 
        scale=1.2,
        text height=0.333cm,
        execute at begin node=\color{black}$\cdots$
      },
    }

    \begin{tikzpicture}[x=1cm, y=.75cm, >=stealth]
        \node [circle, fill=gray!50, minimum size=.5cm] (input) at (0,-2) {$X$};
        \node [rectangle, fill=pink!50, minimum size=0.75cm] (hidden) at (0,0) {$h$};
        \node [circle, fill=black!50, minimum size=.5cm] (output) at (0,2) {$Y$};
        \node at (1,0) (here) {};
        \draw [->] (hidden) to[out=-80, in=-90,looseness=2] (here) to[out=90,in=80,looseness=2] (hidden);
        \draw [->] (input) -- (hidden);
        \draw [->] (hidden) -- (output);
        
        \node[single arrow,draw,single arrow tip angle=120] (output) at (1.7,0) {Unfold};
        
        \node [neuron missing]  at (2.9,0.15) {};
        
        \node at (3.1,0) (a) {};
        \node at (3.9,0) (b) {};
        \draw [->] (a) -- (b);
        
        \node [circle, fill=gray!50, minimum size=.5cm] (input_1) at (4.3,-2) {$x_{t-1}$};
        \node [rectangle, fill=pink!50, minimum size=0.75cm] (hidden_1) at (4.3,0) {$h_{t-1}$};
        \node [circle, fill=black!50, minimum size=.5cm] (output_1) at (4.3,2) {$y_{t-1}$};
        \draw [->] (input_1) -- (hidden_1);
        \draw [->] (hidden_1) -- (output_1);
        
        \node [circle, fill=gray!50, minimum size=1cm] (input_2) at (5.6,-2) {$x_{t}$};
        \node [rectangle, fill=pink!50, minimum size=0.75cm] (hidden_2) at (5.6,0) {$h_{t}$};
        \node [circle, fill=black!50, minimum size=1cm] (output_2) at (5.6,2) {$y_{t}$};
        \draw [->] (input_2) -- (hidden_2);
        \draw [->] (hidden_2) -- (output_2);
        
        \node [circle, fill=gray!50, minimum size=.5cm] (input_3) at (6.9,-2) {$x_{t+1}$};
        \node [rectangle, fill=pink!50, minimum size=0.75cm] (hidden_3) at (6.9,0) {$h_{t+1}$};
        \node [circle, fill=black!50, minimum size=.5cm] (output_3) at (6.9,2) {$y_{t+1}$};
        \draw [->] (input_3) -- (hidden_3);
        \draw [->] (hidden_3) -- (output_3);
        
        \draw [->] (hidden_1) -- (hidden_2);
        \draw [->] (hidden_2) -- (hidden_3);
        
        \node at (7.2,0) (c) {};
        \node at (8.,0) (d) {};
        \draw [->] (c) -- (d);
        
        \node [neuron missing]  at (8.2,0.15) {};
        
    \end{tikzpicture}
    \caption{An illustration of a simple RNN and its unfolded structure through time $t$.}
    \label{fig:RNN}
\end{figure}

A recurrent Neural Network (RNN) is a sequential model that can capture the relationship between items in a sequence. Unlike traditional neural networks, wherein all inputs are independent of each other, RNNs contain artificial neurons with one or more feedback loops. Feedback loops are recurrent cycles over time or sequence, as shown in Figure \ref{fig:RNN}. An established RNN problem is exploding or vanishing gradients. For a long data sequence, the gradient could become increasingly smaller or increasingly larger, which halts the learning. To address this issue, Hochreiter et al. \cite{hochreiter1997long} introduced a long short-term memory (LSTM) model and Cho et al. \cite{cho2014learning} proposed a gated recurrent unit (GRU) model. Both of these networks allow the gradient to flow unchanged in the network, thus preventing exploding or vanishing gradients.

\subsubsection{Siamese Neural Network} \label{sec:snn}

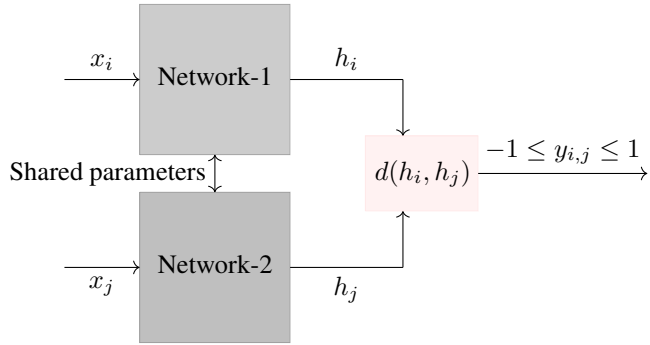
\begin{figure}[h]
	\centering
	\begin{tikzpicture}
	    \draw[fill=black,opacity=0.2,draw=black] (0,0) -- (2,0.0) -- (2,2) -- (0,2) -- (0,0);
	    \draw [<-] (0,1) -- (-1,1) node [above, midway] {$x_{i}$};
		\node at (1,1){\begin{tabular}{c}Network-1 \end{tabular}};
		
		\draw[fill=gray,opacity=0.5,draw=gray] (0,-0.5) -- (2,-0.5) -- (2,-2.5) -- (0,-2.5) -- (0,-.5);
		\draw [<-] (0,-1.5) -- (-1,-1.5) node [below, midway] {$x_{j}$};
		\node at (1,-1.5){\begin{tabular}{c}Network-2 \end{tabular}};
		
		\draw [<->] (1,0) -- (1,-.5) node [left, midway] {Shared parameters};
		
		\draw [-] (2,1) -- (3.5,1) node [above, midway] {$h_i$};
		\draw [->] (3.5,1) -- (3.5,0.25);
		\draw [-] (2,-1.5) -- (3.5,-1.5) node [below, midway] {$h_{j}$};
		\draw [->] (3.5,-1.5) -- (3.5,-0.75) ;
		
		\draw[fill=pink,opacity=0.2,draw=pink] (3,-0.75) -- (4.5,-0.75) -- (4.5,.25) -- (3,.25) -- (3,-.75) node[opacity=1,midway,right]{$d(h_i, h_j)$};
		\draw [->] (4.5,-.25) -- (6.75,-.25) node [above, midway] {$-1 \leq y_{i,j} \leq 1$};
	\end{tikzpicture}
	\caption{An illustration of an SNN. In this figure, $x_i$ and $x_j$ are two data vectors corresponding to a pair of instances from the training set. Both networks share the same weights and map the input to a new representation. By comparing the outputs of the networks using a distance measure such as Euclidean, we can determine the compatibility between instances $x_i$ and $x_j$.}
    \label{fig:SNN}
\end{figure}

There are settings in which the number of training samples is limited, such as in facial recognition scenarios where only one image is available per person. When there is a limited number of examples for each class, DNNs struggle with generalizing the model. One strategy for addressing this problem is to learn a similarity function. This function computes the degree of difference between two samples, instead of learning each class. As an example, let $x_1$ represent one facial image and $x_2$ represent a second. If $d(x1, x2) \leq \tau$, we can conclude that the images are of the same person while $d(x_1, x_2) > \tau$ implies that they are different people. Siamese Neural Networks (SNN) \cite{taigman2014deepface} build on this idea by encoding examples $x_i$ and $x_j$ on two separate DNNs with shared parameters. The SNN learns a function $d$ using encoded features as shown in Figure \ref{fig:SNN}. The network then outputs $y>0$ for similar objects (i.e., when $d$ is less then a threshold value) and $y<0$ otherwise. Thus, SNNs can be used for similarity learning by learning a distance function over objects. In addition to their value for supervised learning from limited samples, SNNs are also beneficial for unsupervised learning tasks \cite{riad2018sampling} \cite{alaverdyan2018regularized}.

\subsection{Unsupervised Learning}

\subsubsection{Generative Adversarial Network} \label{section:gans}

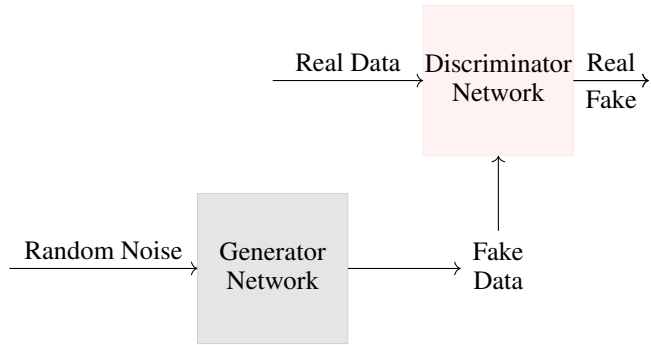
\begin{figure}[h]
	\centering
	\begin{tikzpicture}
	    \draw[fill=pink,opacity=0.2,draw=pink] (3.,0) -- (5.,0.0) -- (5.,2) -- (3.,2) -- (3.,0);
	    \draw [<-] (3,1) -- (1.,1) node [above, midway] {Real Data};
		\node at (4.,1){\begin{tabular}{c}Discriminator \\Network \end{tabular}};
		\draw [->] (5,1) -- (6,1) node [above, midway] {Real};
		\draw [->] (5,1) -- (6,1) node [below, midway] {Fake};
		
		\draw[fill=gray,opacity=0.2,draw=gray] (0,-0.5) -- (2,-0.5) -- (2,-2.5) -- (0,-2.5) -- (0,-.5);
		\draw [<-] (0,-1.5) -- (-2.5,-1.5) node [above, midway] {Random Noise};
		\node at (1,-1.5){\begin{tabular}{c}Generator \\Network \end{tabular}};

		\draw [->] (2,-1.5) -- (3.5,-1.5);
		
		\node at (4,-1.5){\begin{tabular}{c}Fake \\Data \end{tabular}};
		
		\draw [->] (4,-1) -- (4,0);
		
	\end{tikzpicture}
	\caption{An illustration of a GAN. The goal of the discriminator network is to distinguish real data from fake data, and the goal of the generator network is to use the feedback from the discriminator to generate data that the discriminator cannot distinguish from real.}
    \label{fig:GAN}
\end{figure}

Until this point in the survey, we have focused on deep learning for its power in classifying data points. However, researchers have exploited deep learning for other uses as well, such as generating synthetic data that shares characteristics with known real data.
One way to create synthetic data is to learn a generative model. A generative model learns the parameters that govern a distribution based on observation of real data points from that distribution. The learned model can then be used to create arbitrary amounts of synthetic data that emulate the real data observations. Recently, researchers have found a way to exploit multiplayer games for the purpose of improving generative machine learning algorithms. In the adversarial training scenario, two agents compete against each other, as inspired by Samuel \cite{samuel1959some} who designed a computer program to play checkers against itself. Goodfellow et al. \cite{NIPS2014_5423}  put this idea to use in developing Generative Adversarial Networks (GANs), in which a DNN (generator) tries to generate synthetic data that is so similar to real data that it fools its opponent DNN (discriminator), whose job is to distinguish real from fake data (see Figure \ref{fig:GAN} for an illustration). The goal of GANs is to simultaneously improve the ability of the generator to produce realistic data and of the discriminator to distinguish synthetic from real data. GANs have found successful application in diverse tasks including translating text to images \cite{reed2016generative}, discovering drugs \cite{kadurin2017cornucopia}, and transforming sketches to images \cite{chen2018sketchygan} \cite{park2019gaugan}.

\subsubsection{Autoencoder} \label{sec:ae}

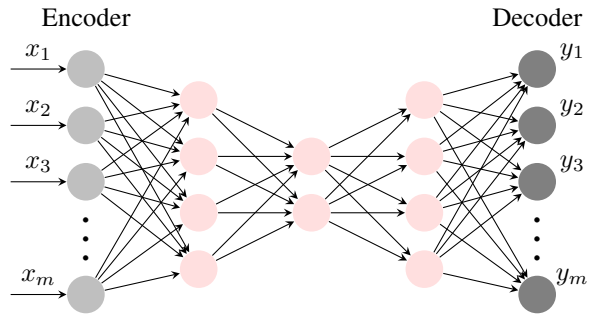
\begin{figure}[h]
   \centering
    \tikzset{%
       neuron missing/.style={
        draw=none, 
        scale=1.8,
        text height=0.333cm,
        execute at begin node=\color{black}$\vdots$
      },
    }
    
    \begin{tikzpicture}[x=1cm, y=.75cm, >=stealth]
    
        \foreach \m/\l [count=\y] in {1,2,3} {
         \node [circle,fill=gray!50,minimum size=.5cm] (input-\m) at (0,2.5-\y) {};
        }
        \foreach \m/\l [count=\y] in {4} {
         \node [circle,fill=gray!50,minimum size=.5cm ] (input-\m) at (0,-2.5) {};
        }
         
        \node [neuron missing]  at (0,-1.5) {};
         
        \foreach \m [count=\y] in {1,...,4}{
          \node [circle,fill=pink!50,minimum size=.5cm ] (hidden1-\m) at (1.5,1.95-\y) {};
        }
        
        \foreach \m [count=\y] in {1,...,2} 
          \node [circle,fill=pink!50,minimum size=.5cm ] (hidden2-\m) at (3,0.95-\y) {};
        
        \foreach \m [count=\y] in {1,...,4}
          \node [circle,fill=pink!50,minimum size=.5cm ] (hidden3-\m) at (4.5,1.95-\y) {};
        
        \foreach \m [count=\y] in {1,2,3}
          \node [circle,fill=black!50,minimum size=.5cm ] (output-\m) at (6,2.5-\y) {};
          
        \foreach \m [count=\y] in {4}
          \node [circle,fill=black!50,minimum size=0.5cm ] (output-\m) at (6,-2.5) {};
        
        \node [neuron missing]  at (6,-1.5) {};
         
        \foreach \l [count=\i] in {1,2,3,m}
          \node [above, xshift=6pt] at (output-\i.east) {$y_{\l}$};
        
        \foreach \l [count=\i] in {1,2,3,m}
          \draw [<-] (input-\i) -- ++(-1,0)
            node [above, midway] {$x_{\l}$};
        
        \foreach \i in {1,...,4}
          \foreach \j in {1,...,4}
            \draw [->] (input-\i) -- (hidden1-\j);
            
        \foreach \i in {1,...,4}
          \foreach \j in {1,...,2}
            \draw [->] (hidden1-\i) -- (hidden2-\j);
            
        \foreach \i in {1,...,2}
          \foreach \j in {1,...,4}
            \draw [->] (hidden2-\i) -- (hidden3-\j);
        
        \foreach \i in {1,...,4}
          \foreach \j in {1,...,4}
            \draw [->] (hidden3-\i) -- (output-\j);
        
        \foreach \l [count=\x from 0] in {Encoder, Decoder}
          \node [align=center, above] at (\x*6,2.1) {\l};
        
    \end{tikzpicture}
    \caption{An illustration of an AE. The first part of the network, called the encoder, compresses input into a latent-space by learning the function $h=f(x)$. The second part, called the decoder, reconstructs the input from the latent-space representation by learning the function $\hat{y}=g(h)$.}
    \label{fig:AE}
\end{figure}

Yet another purpose for deep neural networks is to provide data compression and dimensionality reduction.
An Autoencoder (AE) is a DNN that accomplishes this goal by creating an output layer that resembles the input layer, using a reduced set of terms represented by the middle layers \cite{goodfellow2016deep}. Architecturally, an AE combines two networks. The first network, called the encoder, learns a new representation of input $x$ with fewer features $h=f(x)$; the second part, called the decoder, maps $h$ onto a reconstruction of the input space $\hat{y}=g(h)$, as shown in Figure \ref{fig:AE}. The goal of an AE is not simply to recreate the input features. Instead, an AE learns an approximation of the input features to identify useful properties of the data. AEs are vital tools for dimensionality reduction \cite{hinton2006reducing}, feature learning \cite{vincent2008extracting}, image colorization \cite{zhang2016colorful}, higher-resolution data generation \cite{huang2018image}, and latent space clustering \cite{yeh2017learning}. Additionally, other versions of AEs such as variational autoencoders (VAEs) \cite{kingma2013auto} can be used as generative models.

\subsection{Optimization for Training Deep Neural Networks} \label{sec:sgd}

\begin{figure}[h]
	\centering
	\begin{tikzpicture}
        
        \draw[fill=black!20] (0, 0) circle (0.9);
        \draw[fill=black!40] (0, 0) circle (0.7);
        \draw[fill=black!60] (0, 0) circle (0.5);
        \draw[fill=black!80] (0, 0) circle (0.3);
        \draw[fill=black] (0, 0) circle (0.1);
        
        \draw [->, color=pink, line width=0.5mm] (0.70, 0.6) -- (.4,0.5);
        \draw [->, color=pink, line width=0.5mm] (0.60, -0.6) -- (.3,-0.5);
        \draw [->, color=pink, line width=0.5mm] (0.30, -0.5) -- (.1,-.1);
        \draw [->, color=pink, line width=0.5mm] (0.30, 0.3) -- (.1,0.1);
        \draw [->, color=pink, line width=0.5mm] (-0.40, 0.3) -- (-.0,-0.0);
        \draw [->, color=pink, line width=0.5mm] (-0.50, -0.50) -- (-.3,-0.3);
        \draw [->, color=pink, line width=0.5mm] (-0.30, -0.30) -- (-.1,-0.1);
        \draw [->, color=pink, line width=0.5mm] (-0.40, 0.70) -- (-.4,0.3);

        \draw[fill=black!20] (4, 0) ellipse (1.8 and 0.9);
        \draw[fill=black!40] (4, 0) ellipse (1.4 and 0.7);
        \draw[fill=black!60] (4, 0) ellipse (1 and 0.5); 
        \draw[fill=black!80] (4, 0) ellipse (0.6 and 0.3);
        \draw[fill=black] (4, 0) ellipse (0.2 and 0.1);
        
        \draw [->, color=pink, line width=0.5mm] (2.2, 0) -- (3,0.6);
        \draw [->, color=pink, line width=0.5mm] (5, -.8) -- (5,0.7);
        \draw [->, color=pink, line width=0.5mm] (5.5, 0.3) -- (4,0.8);
        \draw [->, color=pink, line width=0.5mm] (2.5, -0.3) -- (3,-0.3);
        \draw [->, color=pink, line width=0.5mm] (3, -0.3) -- (3.5,0.1);
        \draw [->, color=pink, line width=0.5mm] (3.5,0.1) -- (4,0.0);
	\end{tikzpicture}
	\caption{The left-hand side loss surface depicts a well-conditioned model where local minima can be reached from all directions. The right-hand side loss surface depicts an ill-conditioned model where there are several ways to overshoot or never reach the minima.}
    \label{fig:ill}
\end{figure}
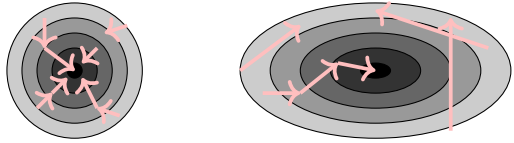

In the previous section, we described common DNN architecture components. In this section, we offer a brief overview of optimization approaches for training DNNs. Learning methods may optimize a function $f(x)$ (e.g., minimize a loss function) by modifying model parameters (e.g., changing DNN weights). However, as Bengio et al. \cite{bengio2013deep} point out, DNN optimization during training may be further complicated by local minima and ill-conditioning (see Figure~\ref{fig:ill} for an illustration of an ill-condition). 

The most common type of optimization strategy employed by DNNs is gradient descent. This intuitive approach to learns the weights of connections between layers which reduce the network's objective function by computing the error derivative with respect to a Ir-level layer of the network. Input $x$ is fed forward through a network to predict $\hat{y}$. A cost function $J(\theta)$ measures the error of the network at the output layer. Gradient descent then directs the cost value to flow backward through the network by computing the gradient of the objective function $\nabla_{\theta}J(\theta)$. This process is sometimes alternatively referred to as backpropagation because the training error propagates backward through the network from output to input layers. Many variations of gradient descent have been tested for DNN optimization, such as  stochastic gradient descent, mini-batch gradient descent, momentum \cite{sutskever2013importance}, Ada-Grad \cite{duchi2011adaptive}, and Adam \cite{kingma2014adam}.

Deep network optimization is an active area of research. Along with gradient descent, many other algorithms such as derivative-free optimization \cite{rios2013derivative} and feedback-alignment \cite{nokland2016direct} have  appeared. However, none of these algorithms are as popular as the gradient descent algorithms.

\subsection{Regularization} \label{sec:regularization}

Regularization was an optimization staple for decades prior to the development of DNNs. The rationale behind adding a regularizer to a classifier is to avoid the overfitting problem, where the classifier fits the training set too closely instead of generalizing to the entire data space. Goodfellow et al. \cite{goodfellow2016deep} defined regularization as ``any modification to a learning algorithm that is intended to reduce its generalization error but not its training error''. While regularization methods such as bagging have been popular for neural networks and other classifiers, recently the DNN community has developed novel regularization methods that are unique to deep neural networks. In some cases, backpropagation training of fully-connected DNNs results in poorer performance than shallow structures because the deeper structure is prone to being trapped in local minima and overfitting the training data. To improve the generalizability of DNNs, regularization methods have thus been adopted during training. Here we review the intuition behind the most frequent regularization methods that are currently found in DNNs.

\subsubsection{Parameter Norm Penalty} \label{sec:penalty}

A conventional method for avoiding overfitting is to penalize large weights by adding a p-norm penalty function to the optimization function of the form  $f(x) +$ {\it p-norm}$(x)$, where the p-norm $p$ for weights $w$ is denoted as $||w||_p = (\sum_i |w_i|^p)^{\frac{1}{p}}$. Popular p-norms are the $L_1$ and $L_2$ norms which have been used by other classifiers such as logistic regression and SVMs prior to the introduction of DNNs. $L_1$ adds a regularization term $\Omega(\theta)=||w||_1$ to the objective function for weights $w$, while $L_2$ adds a regularization term $\Omega(\theta)=||w||_2$. The difference between the $L_1$ and $L_2$ norm penalty functions is that $L_1$ penalizes features more heavily by setting the corresponding edge weights to zero compared to $L_2$. Therefore, a classifier with the $L_1$ norm penalty tends to prefer a sparse model. The $L_2$ norm penalty is more common than the $L_1$ norm penalty. However, it is often advised to use the $L_1$ norm penalty when the amount of training data is small and the number of features is large to avoid noisy and less-important features. Because of its sparsity property, the $L_1$ penalty function is a key component of LASSO feature selection \cite{tibshirani1996regression}.

\subsubsection{Dropout} \label{sec:dropout}

A powerful method to reduce generalization error is to create an ensemble of classifiers. Multiple models are trained separately, then as an ensemble they output a combination of the models' predictions on test points. Some examples of ensemble methods included bagging \cite{breiman1996bagging}, which trains $k$ models on $k$ different folds of random samples with replacement, and boosting \cite{freund1995boosting}, which applies a similar process to weighted data. A variety of DNNs use boosting to achieve lower generalization error \cite{hinton2006fast} \cite{moghimi2016boosted} \cite{eickholt2013dndisorder}. 

Dropout \cite{srivastava2014dropout} is a popular regularization method for DNNs which can be viewed as a computationally-inexpensive application of bagging to deep networks. A common way to apply dropout to a DNN is to deactivate a randomly-selected 50\% of the hidden nodes and a randomly-selected 20\% of the input nodes for each mini-batch of data. The difference between bagging and dropout is that in bagging the models are independent of each other, while in dropout each model inherits a subset of parameters from the parent deep network.

\subsubsection{Data Augmentation} \label{sec:dataaug}

DNNs can generalize better when they have more training data; however, the amount of available data is often limited. One way to circumvent this limitation is to generate artificial data from the same distribution as the training set. Data augmentation has been particularly effective when used in the context of classification. The goal of data augmentation is to generate new training samples from the original training set $(X, y)$ by transforming the $X$ inputs. Data augmentation may include generating noisy data to improve robustness (denoising) or creating additional training data for the purpose of regularization (synthetic data generation). Dataset augmentation has been adopted for a variety of tasks such as image recognition \cite{perez2017effectiveness} \cite{cubuk2018autoaugment}, speech recognition \cite{jaitly2013vocal}, and activity recognition \cite{ohashi2017augmenting}. Additionally, GANs \cite{bowles2018gan} \cite{antoniou2017data} and AEs \cite{jorge2018empirical} \cite{liu2018data}, described in Sections~\ref{section:gans} and~\ref{sec:ae}, can be employed to generate such new examples. 

Injecting noise into a copy of the input is another data augmentation method. Although DNNs are not consistently robust to noise \cite{tang2010deep}, Poole et al. \cite{poole2014analyzing} show that DNNs can benefit from carefully-tuned noise.

\section{Deep Learning Architectures Outside of Deep Neural Networks}

Recent research has introduced numerous enhancements to the basic neural network architecture that enhance network classification power, particularly for deep networks. In this section, we survey non-network classifiers that also make use of these advances.

\subsection{Supervised Learning} 

\subsubsection{Feedforward Learning} \label{sec:outfeedforward}

A DNN involves multiple layers of operations that are performed sequentially. The idea of creating a sequence of operations, each of which manipulates the data before passing them to the next operator, may be used to improve many types of classifiers. One way to construct a model with a deep feedforward architecture is to use stacked generalization \cite{wolpert1992stacked}  \cite{ting1999issues}. Stacked generalization classifiers are comprised of multiple layers of classifiers stacked on top of each other as found in DNNs. In stacked generalization classifiers, one layer generates the next layer's input by concatenating its own input to its output. Stacked generalization classifiers typically only implement forward propagation, in contrast to DNNs which propagate information both forward and backward through the model.

In general, learning methods that employ stacked generalization can be categorized into two strategies. In the first stacked generalization strategy, the new feature space for the current layer comes from the concatenation of the predicted output of the previous layer with the original feature vector. Here, layers refer not to layers of neural network operations, but instead refer to sequences of other types of operations. Examples of this strategy include Deep Forest (DF) \cite{zhou2017deep} and the Deep Transfer Additive Kernel Least Square SVM (DTA-LS-SVM) \cite{wang2017deep}. At any given layer, for each instance $x$, DF extends $x$'s previous feature vector to include the previous layer's predicted class value for the instance. The prediction represents a distribution over class values, averaged over all trees in the forest. Furthermore, Zhou et al. \cite{zhou2017deep} introduce a method called Multi-Grained Scanning for improving the accuracy of DFs. Inspired by CNNs and RNNs where spatial relationships between features are critical, Multi-Grained Scanning splits a $D$-dimensional feature vector into smaller segments by moving a window by moving a window over the features. For example, given 400 features and a window size of 100, the original features convert to 301 features of length 100,  $\{<1-100>, <2-101>, \ldots, <301-400>\}$, where the new instances have the same labels as the original instances. The new samples which are described by a subset of the original features might have incorrectly-associated labels. At a first glance, it seems these noisy data could hurt the generalization. But as Breiman illustrates \cite{breiman2000randomizing}, perturbing a percentage of the training labels can actually help generalization. 

Furthermore, Ho \cite{ho1995random} demonstrates that feature sub-sampling can enhance the generalization capability for RFs. Zhou et al. \cite{zhou2017deep} tested three different window sizes ($D/4$, $D/8$, and $D/16$), where data from each different window size fits a different level of a DF model. Then the newly-learned representation from these three layers are fed to a multilayer DF, applying subsampling when the transformed features are too long.  Multi-Grained Scanning can improve the performance of a DF model for continuous data, as Zhou et al. \cite{zhou2017deep} report that accuracy increased by 1.24\% on the MNIST \cite{lecun1998mnist} dataset. An alternative method, DTA-LS-SVM, applies an Additive Kernel Least Squares SVM (AK-LS-SVM) \cite{cawley2006leave} \cite{yang2012practical} at each layer and concatenates the original feature vector $x$ with the prediction of the previous level to feed to the next layer. In addition, DTA-LS-SVM incorporates a parameter-transfer approach between the source (previous-layer learner) and target (next-layer learner) to enhance the classification capability of the higher level.

In the second stacked generalization strategy, the current layer's new feature space comes from the concatenation of predictions from all previous layers with the original input feature vector. Examples of this strategy include the Deep SVM (D-SVM) \cite{abdullah2009ensemble} and the Random Recursive SVM (R2-SVM) \cite{vinyals2012learning}. The D-SVM contains multiple layers of SVMs, where the first layer is trained in the normal fashion. Following this step, each successive layer employs the kernel activation from the previous layer with the desired labels. The R2-SVM is a multilayer SVM model which at each layer transforms the data based on the sigmoid of a projection of all previous layers' outputs. For the data $(X,Y)$ where $X \in R^D$ and $Y \in R^C$, the random projection matrix is $W \in R^{D \times C}$, where each element is sampled from $N(0, 1)$. The input data for the next layer is:
\begin{equation} \label{eq:R2SVM}
X_{l+1} = \sigma(d+\beta W_{l+1}[o_{1}^{T}, o_{2}^{T}, ..., o_{l}^{T}]^T),
\end{equation}
\noindent where $\beta$ is a weight parameter that controls the degree with which a data sample in $X_{l+1}$ moves from the previous layer, $\sigma(.)$ is the sigmoid function, $W_{l+1}$ is the concatenation of $l$ random projection matrices $[W_{l+1,1}, W_{l+1,2}, ... ,W_{l+1,l}]$, one for each previous layer, and $o$ is the output of each layer. Addition of a sigmoid function to the recursive model prevents deterioration to a trivial linear model in a similar fashion as MLPs. The purpose of the random projection is to push data from different classes in different directions.

It is important to note here that stacked generalization can be found in DNNs as well as non-network classifiers. Examples of DNNs with stacked generalization include Deep Stacking Networks \cite{deng2012scalable} \cite{hutchinson2013tensor} and Convex Stacking Architectures \cite{deng2011deep} \cite{deng2012scalable}. This is clearly one enhancement that benefits all types of classifier strategies. However, there is no evidence that stack generalization could add nonlinearity to the model.

DNN classifiers learn a new representation of data at each layer with a goal that the newly-learned representation maximally separate the classes. Unsupervised DNNs often share this goal. As an example, Deep PCA's model \cite{liong2013face} is made of two layers that each learn a new data representation by applying a Zero Components Analysis (ZCA) whitening filter \cite{krizhevsky2009learning} followed by a principal components analysis (PCA) \cite{shlens2014tutorial}. The final data representation is derived from concatenating the output of the two layers. The motivation behind applying a ZCA whitening filter is to force the model to focus on higher-order correlations. One motivation for combining output from the first and second layers could be to preserve the learned representation from the first layer and to prevent feature loss after applying PCA at each layer. Experiments demonstrate that Deep PCA exhibits superior performance for face recognition tasks compared to standard PCA and a two-layer PCA without a whitening filter. However, as empirically confirmed by Damianou et al. \cite{damianou2013deep}, stacking PCAs does not necessarily result in an improved representation of the data because Deep PCA is unable to learn a nonlinear representation of data at each layer. Damianou et al. \cite{damianou2013deep} fed a Gaussian to a Deep PCA and observed that the model learned just a lower rank of the input Gaussian at each layer.

As pointed out earlier in this survey, the invention of the deep belief net (DBN) \cite{hinton2006fast} drew the attention of researchers to developing deep models. A DBN can be viewed as a stacked restricted Boltzmann machine (RBM), where each layer is trained separately and alternates functionality between hidden and input units. In this model, features learned at hidden layers then represent inputs to the next layer. A RBM is a generative model that contains a single hidden layer. Unlike the Boltzmann machine, hidden units in the restricted model are not connected to each other and contain undirected, symmetrical connections from a layer of visible units (inputs). All of the units in each layer of a RBM are updated in parallel by inputting the current state of the unit to the other layer. This updating process repeats until the system is sampling from an equilibrium distribution. The RBM learning rule is shown in Equation~\ref{eq:rbm}.
\begin{equation}
   \label{eq:rbm}
    \frac{\partial \log P(v)}{\partial W_{ij}} \approx <v_i h_j>_{data} - <v_i h_j>_{reconstruction}
\end{equation}
In this equation, $W_{ij}$ represents the weight vector between a visible unit $v_{i}$ and a hidden unit $h_{j}$, and $<.>$ is the average value over all training samples. Since the introduction of DBNs, many other different variations of Deep RBMs have been proposed such as temporal RBMs \cite{sutskever2007learning}, gated RBMs \cite{memisevic2007unsupervised}, and cardinality RBMs \cite{swersky2012cardinality}.

Another novel form of a deep belief net is a deep Gaussian process (DGP) model \cite{damianou2013deep}. DGP is a deep directed graph where multiple layers of Gaussian processes map the original features to a series of latent spaces. DGPs offer a more general form of Gaussian Processes (GPs) \cite{rasmussen2003gaussian} where a one-layer DGP consists of a single GP, $f$. In a multilayer DGP, each GP, $f_l$, maps data from one latent space to the next. As shown in Equation~\ref{eq:dgp}, each data point $Y$ is generated from the corresponding function $f_l$ with $\epsilon$ Guassian noise applied to data $X_l$ that is obtained from a previous layer.
\begin{equation}
    \label{eq:dgp}
    Y = f_l(X_l)+\epsilon_l, \indent \epsilon_l \sim \mathcal{N}(0, \sigma^{2}_{l}I)
\end{equation}

\begin{figure}[h]
   \centering
    \tikzset{%
       neuron missing/.style={
        draw=none, 
        scale=1.8,
        text height=0.333cm,
        execute at begin node=\color{black}$\vdots$
      },
    }
    
    \begin{tikzpicture}[x=1cm, y=.75cm, >=stealth]
        \node [circle,fill=pink!50,minimum size=.75cm] (input) at (0,2.5) {};
        \node [circle,fill=pink!50,minimum size=.75cm] (hidden1) at (2,2.5) {};
        \node [circle,fill=pink!50,minimum size=.75cm] (hidden2) at (4,2.5) {};
        \node [circle,fill=gray!50,minimum size=0.75cm ] (output) at (6,2.5) {};
        \draw [->] (input) -- (hidden1);
        \draw [->] (hidden1) -- (hidden2);
        \draw [->] (hidden2) -- (output);
        \draw [{Circle[black]}-latex] (-2,2.5)node[left] {X} -- (-.30,2.5 ) ;
        \node [] at (input) {$f_1$};
        \node [] at (hidden1) {$f_2$};
        \node [] at (hidden2) {$f_3$};
        \node [] at (output) {$Y$};

    \end{tikzpicture}
    \caption{A deep Gaussian process with two hidden layers.}
    \label{fig:DGP}
\end{figure}
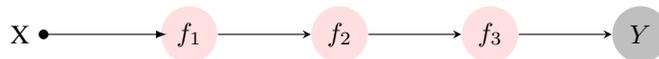

Figure \ref{fig:DGP} illustrates a DGP expressed as a series of Gaussian processes mapping data from one latent space to the next. Functions $f_l$ are drawn from a Gaussian process, i.e. $f(x) \sim \mathcal{GP}(0, k(x,x'))$. In this setting, the covariance function $k$ defines the properties of the mapping function. DGP can be utilized for both supervised and unsupervised learning. In the supervised setting, the top hidden layer is observed, whereas in the unsupervised setting, the top hidden layer is set to a unit Gaussian as a fairly uninformative prior. DGP is a powerful non-parametric model but it has only been tested on small datasets. Also, we note that researchers have developed deep Gaussian process models with alternative architectures such as recurrent Gaussian processes \cite{mattos2015recurrent}, convolutional Gaussian processes \cite{van2017convolutional} and variational auto-encoded deep Gaussian processes \cite{dai2015variational}. There exists a vast amount of literature on this topic that provides additional insights on deep Gaussian processes \cite{damianou2015deep} \cite{dunlop2018deep} \cite{duvenaud2014avoiding}.

As we discussed, non-network classifiers have been designed that contain multiple layers of operations, similar to a DNN. We observe that a common strategy for creating a deep non-network model is to add the prediction of the previous layer or layers to the original input feature. Likewise, novel methods can be applied to learn a new representation of data at each layer. We discuss these methods next.

\subsubsection{Siamese Model}

As discussed in Section \ref{sec:snn}, a SNN represents a powerful method for similarity learning. However, one problem with SNNs is overfitting when there is a small number of training examples. The Siamese Deep Forest (SDF) \cite{utkin2017siamese} is a method based on DF which offers an alternative to a standard SNN. The SDF, unlike the SNN, uses only one DF. The first step in training a SDF is to modify the training examples. The training set consists of the concatenation of each pair of samples in the original set. If sample points $x_i$ and $x_j$ are semantically similar, the corresponding class label is set to zero;  otherwise, the class label is set to one. The difference between the SDF and the DF in training is that the Siamese Deep Forest concatenates the original feature vector with a weighted sum of the tree class probabilities. Training of SDF is similar to DF; the primary difference is that SDF learns the class probability weights $w$ for each forest separately at each layer. 
Learning the weights for each forest can be accomplished by minimizing the function in Equation~\ref{eq3}.
\begin{equation} \label{eq3}
\min\limits_{w}J_{q}(w) = \min\limits_{w} \sum_{i,j}l(x_i, x_j, y_{ij}, w) + \lambda R(w)
\end{equation}
Here, $w$ represents a concatenation of vectors $w^{k}$, $k= 1, ..., M$, $q$ is the SDF layer, $R(w)$ is a regularization term, and $\lambda$ is a hyper-parameter to control regularization. Detailed instructions on minimizing Equation \ref{eq3} are found in the literature  \cite{utkin2017siamese}. The results of SDF experiments indicate that the SDF can achieve better classification accuracy than DF for small datasets. In general, all non-network models that learn data representations can take advantage of the Siamese architecture like SDF.

\subsection{Unsupervised Learning}

\subsubsection{Generative Adversarial Model}

A common element found in GANs is inclusion of a FC layer in the discriminator. One issue with the FC layer is that it cannot deal with the ill-condition in which local minima are not surrounded by spherical wells as shown in Figure \ref{fig:ill}. The Generative Adversarial Forest (GAF) \cite{zuo2018generative} replaces the FC layer of the discriminator with a deep neural decision forest (DNDF), which is discussed in Section \ref{sec:optoutofnet}. GAF and DNDF are distinguished based on how leaf node values are learned. Instead of learning leaf node values iteratively, as DNDF does, GAF learns them in parallel across the ensemble members. The strong discriminatory power of the decision forest is the reason the authors recommend this method in lieu of the fully-connected discriminator layer. 

In this previous work, the discriminator is replaced by an unconventional model. We hypothesize that replacing the discriminator with other classifiers such as Random Forest, SVM, of K nearest neighbor based on the data could result in a diverse GAN strategies, each of which may offer benefits for alternative learning problems.

\subsubsection{Autoencoder}

As we discussed in Section \ref{sec:ae}, AEs offer strategies for dimensionality reduction and data reconstruction from compressed information. The autoencoding methodology can be found in neural networks, non-networks, and hybrid methods. As an example, the multilayer SVM (ML-SVM) autoencoder is a variation of ML-SVM with the same number of output nodes as input features and a single hidden layer that consists of fewer nodes than the input features. ML-SVM is a model with the same structure as a MLP. The distinction here is that the network contains SVM models as its nodes. A review of ML-SVM is discussed in Section \ref{sec:optoutofnet}. The outputs of hidden nodes are fed as input to each SVM output node $c$ as follows:
\begin{equation} \label{eq:ML_SVM_AE}
g_c(f(X|\theta)) = \sum_{i=1}^{l}(\alpha_{i}^{c*} - \alpha_{i}^{c})K_{o} (f(x_i|\theta),f(x|\theta)) + b_c,
\end{equation}
\noindent where $\alpha_{i}^{c*}$ and $\alpha_{i}^{c}$ are the support vector coefficients, $K_o$ is the kernel function, and $b_c$ is their bias. The error backpropagates through the network to update the parameters.

Another exciting emerging research area is the combination of Kalman filters with deep networks. A Kalman filter is a well-known algorithm that estimates the optimal state of a system from a series of noisy observations. The classical Kalman filter \cite{kalman1960new} is a linear dynamical system and therefore is unable to model complex phenomena. For this reason, researchers developed nonlinear versions of Kalman filters. In a seminal contribution,  Krishnan et al. \cite{Krishnan2015kalman} introduced a model that combines a variational autoencoder with Kalman filters for counterfactual inference of patient information. In a standard autoencoder, the model learns a latent space that represents the original data minus extraneous information or ``signal noise''. In contrast, a variational autoencoder (VAE) \cite{kingma2013auto} adds a constraint to the encoder that it learn a Gaussian distribution of the original input data. Therefore, a VAE is able to generate a latent vector by sampling from the learned Gaussian distribution.
Deep Kalman filters (DKF) learn a generative model from observed sequences $\vec{x} = (x_1, \cdots, x_T)$ and actions $\vec{u} = (u_1, \cdots u_{T-1})$, with a corresponding latent space $\vec{z} = (z_1, \cdots, z_T)$, as follows:
\begin{equation}
    \begin{split}
    & z_1 \sim \mathcal{N}(\mu_0, \Sigma_0) \\
    & z_t \sim \mathcal{N}(G_{\alpha}(z_{t-1},u_{t-1},\Delta_t), S_{\beta}(z_{t-1}, y_{t-1}, \Delta_t)) \\
    & x_t \sim \Pi(F_k(z_t)), \\
    \end{split}
\end{equation}
where $\mu_0 = 0$ and $\Sigma_0 = I_d$, $\Delta_t$ represents the difference between times $t$ and $t-1$, and $\Pi$ represents a distribution (e.g., Bernoulli for binary data) over observation $x_t$. The functions $G_\alpha$, $S_\beta$, $F_k$ are parameterized by a neural net. As a result, the autoencoder will learn $\theta=\{\alpha, \beta, k\}$ parameters. Additionally, Shashua et al. \cite{shashua2017deep} introduced  deep Q-learning with Kalman filters and Lu et al. \cite{lu2018deep} presented a deep Kalman filter model for video compression.

As we highlighted in this section, non-network methods have been designed that are inspired by AEs. Although ML-SVM mimics the architecture of AEs, its computational cost prevents the algorithm from being a practical choice. DKF takes advantage of the VAE idea by learning a Kalman Filter in its middle layer. Additionally, Feng et al. \cite{feng2017autoencoder} introduced an encoder forest, a model inspired by the DNN autoencoder. Because the encoder forest is not a deep model, we do not include the details of this algorithm in our survey.

\section{Deep Learning Optimization Outside of Deep Neural Networks} \label{sec:optoutofnet}

As discussed in Section \ref{sec:sgd}, gradient descent has been a prominent optimization algorithm for DNNs; however, it has been underutilized by non-network classifiers. Some notable exceptions are found in the literature. We discuss these here.

A resourceful method for constructing a deep model is to start with a DNN architecture and then replace nodes with non-network classifiers. As an example, the multilayer SVM (ML-SVM) \cite{wiering2014multi} replaces nodes in a MLP with standard SVMs. ML-SVM is a multiclass classifier which contains SVMs within the network. At the output layer, the ML-SVM contains the same number of SVMs as the number of classes learned at the perceptron output layer. Each SVM at the ML-SVM output layer is trained in a one-versus-all fashion for one of the classes. When observing a new data point, ML-SVM outputs the class label corresponding to the SVM that generates the highest confidence. At each hidden layer, SVMs are associated with each node that learn latent variables. These variables are then fed to the output layer. At hidden layer $f(X|\theta)$ where $X$ is the training set and $\theta$ denotes the trainable parameters of the SVM, ML-SVM maps the hidden layer features to an output value as follows: 
\begin{equation} \label{eq:ML-SVM}
g(f(X|\theta)) = \sum^{l}_{i=1}y^{c}_{i} \alpha^{c}_{i} K_{o}(f(x_i|\theta), f(X|\theta))+b_c,
\end{equation}
\noindent where $g$ is the output layer function, $y_i^c \in \{-1, 1\}$ for each class $c$, $K_o$ is the kernel function for the output layer, $\alpha_i^c$ are the support vector coefficients for SVM nodes of the output layer, and $b_c$ is their bias. The goal of ML-SVM is to learn the maximum support vector coefficient of each SVM at the output layer with respect to the objective function $J_c(.)$, as shown in Equation~\ref{eq:ML-SVM_objfunc}.
\begin{equation} \label{eq:ML-SVM_objfunc}
\min_{w^{c}, b, \xi, \theta} J_c = \frac{1}{2} ||w^{c}||^{2} + C \sum_{i}^{l} \xi_{i}
\end{equation}
\noindent Here, $w^{c}$ represents the set of weights for class $c$, $C$ represents a trade-off between margin width and misclassification risk and $\xi_{i}$ are slack variables. ML-SVM applies gradient ascent to adapt its support vector coefficient towards a local maximum of $J_c(.)$. The support vector coefficient is defined as zero for values less than zero and is assigned to $C$ for values larger than $C$. The data is backpropagated through the network similar to traditional MLPs by calculating the gradient of the objective function. 

The SVMs in the hidden layer are identical. Given the same inputs, they would thus generate the same outputs. To diversity the SVMs, the hidden layers train on a perturbed version of the training set to eliminate producing similar outputs before training the combined ML-SVM model. The outputs of hidden layer nodes are constrained to generate values in the range $[-1 \ldots 1]$. Despite the effort of ML-SVMs to learn a multi-layer data representation, this approach is currently not practical because adding a new node incurs a dramatic computational expense for large datasets.

Kontschieder et al. \cite{kontschieder2015deep} further incorporate gradient descent into a Random Forest (RF), which is a popular classification method. One of the drawbacks of a RF is that does not traditionally learn new internal representations like DNNs. The Deep Network Decision Forest (DNDF) \cite{kontschieder2015deep} integrates a DNN into each decision tree within the forest to reduce the uncertainty at each decision node. In DNDF, the result of a decision node $d_n(x, \Theta)$ corresponds to the output of a DNN $f_n(x, \Theta)$, where $x$ is an input and $\Theta$ is the parameter of a decision node.  DNDF must have differentiable decision trees to be able to apply gradient descent to the process of updating decision nodes. In a standard decision tree, the result of a decision node $d_n(x, \Theta)$ is deterministic. DNDF replaces the traditional decision node with a sigmoid function $d_n(x, \Theta) = \sigma(f_n(x;\Theta))$ to create a stochastic decision node. The probability of reaching a leaf node $l$ is calculated as the product of all decision node outputs from the root to the leaf $l$, which is expressed as $\mu_l$ in this context. The set of leaf nodes $\mathcal{L}$ learns the class distribution $\pi$, and the class with the highest probability is the prediction of the tree. The aim of DNDF is to minimize its empirical risk with respect to the decision node parameter $\Theta$ and the class distribution $\pi$ of $\mathcal{L}$ under the log-loss function for a given data set.

The optimization of the empirical risk is a two-step process which is executed iteratively. The first step is to optimize the class distribution of leaf nodes $\pi_\mathcal{L}$ while fixing the decision node parameters and the corresponding DNN. At the start of optimization (iteration 0), class distribution $\pi^{{0}}$ is set to a uniform distribution across all leaves. DNDF then iteratively updates the class distribution across the leaf nodes as follows for iteration t+1:
\begin{equation} \label{pred}
\pi_{l_{y}}^{(t+1)} = \frac{1}{Z_{l}^{(t)}} \sum_{(x,y')\in \mathcal{T}} \frac{\mathbbm{1}_{y=y'} \pi_{l_{y}^{(t)}}\mu_{l}(x|\Theta)}{\mathbb{P}_{T}[y|x,\Theta, \pi^{(t)}]},
\end{equation}
\noindent where $Z_{l}^{(t)}$ is a normalization factor ensuring that $\sum_{y} \pi^{t+1}_{l_y} = 1$, $\mathbbm{1}_q$ is the indicator function on the argument $q$, and $\mathbb{P}_{T}$ is the prediction of the tree.

The second step is to optimize decision node parameters $\Theta$ while fixing the class distribution $\pi_{\mathcal{L}}$. DNDF employs gradient descent to minimize log-loss with respect to $\Theta$ as follows:
\begin{equation} \label{dis}
\frac{\partial L}{\partial \Theta}(\Theta, \pi; x, y) = \sum_{n \in \mathcal{N}} \frac{\partial L(\Theta,\pi;x,y)}{\partial f_n(x;\Theta)} \frac{\partial f_n(x;\Theta)}{\partial \Theta}.
\end{equation}
The second term in Equation \ref{dis} is the gradient of the DNN. Because this is commonly known, we only discuss calculating the gradient of the differentiable decision tree. Here, the gradient of the differentiable decision tree is given by:
\begin{equation} \label{dt}
\frac{\partial L(\Theta,\pi;x,y)}{\partial f_n(x;\Theta)} = d_n(x; \Theta)A_{n_{r}} - \bar{d}_n(x; \Theta)A_{n_{l}}, 
\end{equation}
\noindent where $d_n$ is the probability of transitioning to the left child, $\bar{d}_n = 1 - d_n$ is the probability of transitioning to the right child calculated by a forward pass through the DNN, and $n_l$ and $n_r$ indicate the left and right children of node $n$. To calculate the term $A$ in Equation \ref{dt}, DNDF performs one forward pass and one backward pass through the differentiable decision tree. Upon completing the forward pass, a value $A_l$ can be initially computed for each leaf node as follows:
\begin{equation} \label{A}
A_{l} = \frac{\pi_{l_{y}}\mu_{l}}{\sum_{l}\pi_{l_{y}} \mu_{l}}.
\end{equation}
\noindent Next, the values of $A_l$ for each leaf node are used to compute the values of $A_m$ for each internal node $m$. To do this, a backward pass is made through the decision tree, during which the values are calculated as $A_m = A_{n_{l}} + A_{n_{r}}$, where $n_l$ and $n_r$ represent the left and the right children of node $m$, respectively.

Each layer of a standard DNN produces the output $o_i$ at layer $i$. As mentioned earlier, the goal of the DNN is to learn a mapping function $F_i:  o_{i-1} \rightarrow o_i$ that minimizes the empirical loss at the last layer of DNN on a training set. Because each $F_i$ is differentiable, a DNN updates its parameters efficiently by applying gradient descent to reduce the empirical loss.

Adopting a different methodology, Frosst et al. \cite{frosst2017distilling} distill a neural network into a soft decision tree. This model benefits from both neural network-based representation learning and decision tree-based concept explainability. In the Frosst soft decision tree (FSDT), each tree's inner node learns a filter $w_i$ and a bias $b_i$, and leaf nodes $l$ learn a distribution of classes. Like the hidden units of a neural network, each inner node of the tree determines the probability of input $x$ at node $i$ as follows:
\begin{equation} 
p_i(x) = \sigma(\beta(xw_i + b_i))
\end{equation}
where $\sigma$ represents the sigmoid function and $\beta$ represents an inverse temperature whose function is to avoid soft decisions in the tree. Filter activation routes the sample $x$ to the left branch for values of $p_i$ less than $0.5$, and  to the right branch otherwise. The probability distribution $Q^{l}$ for each leaf node $l$ represents the learned parameter $\phi^{l}$ at that leaf over the possible $k$ output classes:
\begin{equation}
    Q_{k}^{l}=\frac{\exp(\phi_{k}^{l})}{\sum_{k'}\exp(\phi_{k'}^{l})}.
\end{equation}
The predictive distribution over classes is calculated by traversing the greatest-probability path. To train this soft decision tree, Frosst et al. \cite{frosst2017distilling} calculate a loss function $L$ that minimizes the cross entropy between each leaf, weighted by input vector $x$ path probability and target distribution $T$, as follows: 
\begin{equation}
    L(x)=-\log\Big(\sum_{l \in Leaf Nodes} P^{l}(x)\sum_{k}T_{k}\log Q_{k}^{l}\Big)
\end{equation}    
where $P^{l}(x)$ is the probability of reaching leaf node $l$ given input $x$. Frosst et al. \cite{frosst2017distilling} also introduce a regularization term to avoid internal nodes routing all data points on one particular path and encourage them to equally route data along the left and right branches. The penalty function calculates a sum over all internal nodes from the root to node $i$, as follows:
\begin{equation}
    C = -\lambda \sum_{i \in Inner Nodes} 0.5 \log(\alpha_{i}) + 0.5 \log(1-\alpha_{i})
\end{equation}
where $\lambda$ is a hyper-parameter set prior to training to determine the effect of the penalty. The cross entropy $\alpha$ for a node $i$ is the sum of the path probability $P^{i}(x)$ from the root to node $i$ multiplied by the probability of that node $p_i$ divided by the path probability, as follows:
\begin{equation}
    \alpha_i = \frac{\sum_{x}P^{i}(x)p_{i}(x)}{\sum_{x}P^{i}(x)}.
\end{equation}
Because the probability distribution is not uniform across nodes in the penultimate level, this penalty function could actually hurt the generalization. The authors address this problem by decaying the strength of penalty function $\lambda$ exponentially with the depth $d$ of the node to $2^{d}$. Another challenge is that in any given batch of data, as the data descends the tree, the number of samples decreases exponentially. Therefore, the estimated probability loses accuracy further down the tree. Frosst et al. recommend addressing this problem by decaying a running average of the actual probabilities with a time window that is exponentially proportional to the depth of the nodes \cite{frosst2017distilling}. Although the authors report that the accuracy of this model was less than the deep neural network, the model offers an advantage of concept interpretability.

Both DNDF and the soft decision tree fix the depth of the learned tree to a predefined value. In contrast, Tanno et al. \cite{tanno2018adaptive} introduced the Adaptive Neural Tree (ANT) which can grow to any arbitrary depth. The ANT architecture is similar to a decision tree but at each internal node and edge, ANT learns a new data representation. For example, an ANT may contain one or more convolution layers followed by a fully-connected layer at each inner node, one or more convolution layers followed by an activation function such as \textit{ReLU} or \textit{tanh} at each edge, and a linear classifier at each leaf node.

Training an ANT requires two phases: growth and refinement. In the growth phase, starting from the root in breadth-first order one of the nodes is selected. The learner then evaluates three choices: 1) split the node and add a sub-tree, 2) deepen edge transformation by adding another layer of convolution, or 3) keep the current model. The model optimizes the parameters of the newly-added components by minimizing log likelihood via gradient descent while fixing the parameters of the previous portion of the tree. Eventually, the model selects the choice that yields the lowest log likelihood. This process repeats until the model converges. In the refinement phase, the model performs gradient descent on the final architecture. The purpose of the refinement phase is to correct suboptimal decisions that may have occurred during the growth phase. The authors evaluate their method on several standard testbeds and the results indicate that ANT is competitive with many deep network and non-network learners for these tasks.

Yang et al. \cite{yang2018deep} took a different approach; instead of integrating artificial neurons into the tree, they obtained a decision tree using a neural network. The Deep Neural Decision Tree (DNDT) employs a soft binning function to learn the split rules of the tree. DNDT construct a one-layer neural network with softmax as its activation function. The objective function of this network is:
\begin{equation}
    \text{softmax} \big{(} \frac{wx+b}{\tau} \big{)}.
\end{equation}{}
Here, for a continuous variable $x$, we want to bin it to $n+1$, $w = [1,2, \cdots ,n+1]$ is an untrainable constant, $b$ is a learnable bin or the cutting rule in the tree, and $\tau$ is a temperature variable. After training this model, the decision tree is constructed via the Kronecker product $\otimes$. Given an input $x\in R^D$ with $D$ features, the tree rule to reach a leaf node is:
\begin{equation}
    z = f_1(x_1) \otimes f_2(x_2) \otimes \cdots \otimes f_D(x_D) 
\end{equation}{}
Here, $z$ is an almost-one-hot encoded vector that indicates the index of leaf node. One of the shortcomings of this method is that it cannot handle a high-dimensional dataset because the cost of calculating the Kronecker product becomes prohibitive. To overcome this problem, authors learn a classifier forest by training each tree on a random subset of features.

In some cases, the mapping function is not differentiable. Feng et al. \cite{feng2018multi} propose a new learning paradigm for training a multilayer Gradient Boosting decision tree (mGBDT) \cite{feng2018multi} where $F_i$ is not differentiable. Gradient boosting decision tree (GBDT) is an iterative method which learns an ensemble of regression predictors. In GBDT, a decision tree first learns a model on a training set, then it computes the corresponding error residual for each training sample. A new tree learns a model on the error residuals, and by combining these two trees GBDT is able to learn a more complex model. The algorithm follows this procedure iteratively until it meets a prespecified number of trees for training.

Since gradient descent is not applicable to mGBDT, Feng et al. \cite{feng2018multi} obtain a ``pseudo-inverse'' mapping. In this mapping, $G^{t}_{i}$ represents the pseudo-inverse of  $F^{t-1}_{i}$ at iteration $t$, such that $G^{t}_{i}(F^{t-1}_{i}(o_{i-1})) \sim o_{i-1}$. After performing backward propagation and calculating $G^{t}_{i}$, forward propagation is performed by fitting a pseudo-label $z^{t}_{i-1}$ from $G^{t}_{i}$ to $F^{t-1}_{i}$. The last layer $F_m$ computes $z^{t}_{m}$ based on the true labels at iteration $t$, where $i \in \{2 \ldots m\}$. After this step, pseudo-labels for previous layers are computed via pseudo-inverse mapping. To initialize mGBDT at iteration $t = 0$, each intermediate (hidden) layer outputs Gaussian noise and $F^{0}_{i}$ represent depth-constrained trees that will later be refined. Feng et al. \cite{feng2018multi} thus create a method that is inspired by gradient descent yet is applicable in situations where true gradient descent cannot be effectively applied.

In this section, we examine methods that apply gradient descent to non-network models. As we observed, one way of utilizing gradient descent is to replace the hidden units in a network with a differentiable algorithm like SVM. Another common theme we recognized was to transform deterministic decision-tree nodes into stochastic versions that offer greater representational power. Alternatively, trees or other ruled-based models can be built using neural networks.

\section{Deep Learning Regularization Outside of Deep Neural Networks} \label{sec:regoutdnn}

We have discussed some of the common regularization methods used by DNNs in Section \ref{sec:regularization}. Now we focus on how these methods have been applied to non-network classifiers in the literature. It is worth mentioning that while most models introduced in this section are not deep models, we investigate how non-network models can improve their performance by applying regularization methods typically associated with the deep operations found in DNNs.

\subsection{Parameter Norm Penalty}

Problems arise when a model is learned from data that contain a large number of redundant features. For example, selecting relevant genes associated with different types of cancer is challenging because of a large number of redundancies may exist in the gene's long string of features. There are two common ways to eliminate redundant features: the first way is to perform feature selection and then train a classifier from the selected features; the second way is to simultaneously perform feature selection and classification. As we discussed in Section \ref{sec:penalty}, DNNs apply a $L_1$ or $L_2$ penalty function to penalize large weights. In this section, we investigate how the traditional DNN idea of penalizing features can be applied to non-network classifiers to simultaneously select high-ranked features and perform classification. 

Standard SVMs employ the $L_2$ norm penalty to penalize weights in a manner similar to DNNs. However, the Newton Linear Programming SVM (NLP-SVM) \cite{fung2004feature} replaces the $L_2$ norm penalty with the $L_1$ norm penalty. This has the effect of setting small hyperparameter coefficients to zero, thus enabling NLP-SVM to select important features automatically. A different way to penalize non-important features in SVMs is to employ a Smoothly Clipped Absolute Deviation (SCAD) \cite{zhang2005gene} function. The $L_1$ penalty function can be biased because it imposes a larger penalty on large coefficients; in contrast, SCAD can give a nearly unbiased estimation of large coefficients. SCAD learns a non-convex penalty function as shown in Equation~\ref{eq:scad}.
\begin{equation} \label{eq:scad} 
  p_{\lambda}(|w|)=\begin{cases}
    \lambda|w|  & \text{if } |w| \leq \lambda \\
    -\frac{(|w|^2 - 2a\lambda|w|+\lambda^2)}{2(a-1)}  & \text{if } \lambda < |w| \leq a \lambda \\
    \frac{(a+1)\lambda^2}{2}  & \text{if } |w| > a\lambda
  \end{cases}
\end{equation}
SCAD equates with $L_1$ penalty function until $|w| = \lambda$, then smoothly  transitions to a quadratic function until $|w| = a\lambda$, after which it remains a constant for all $|w|> a\lambda$. As shown by Fan et al. \cite{fan2001variable}, SCAD has better theoretical properties than the $L_1$ function.

One limitation of decision tree classifiers is that the number of training instances that can be selected at each branch in the tree decreases with the tree depth. This downward sampling may cause less relevant or redundant features to be selected near the bottom of the tree. To address this issue, Dang et al. \cite{deng2012feature}  proposed to penalize features that were never selected in the process of making a tree. In a Regularized Random Forest (RRF) \cite{deng2012feature}, the information gain for a feature $j$ is specified as follows:
\begin{equation} \label{12}
  Gain(j)=\begin{cases}
    \lambda . Gain(j)  & \text{$j \not \in F$} \\
    Gain(f_i) & \text{$j \in F$} 
  \end{cases}
\end{equation}
\noindent where $F$ is the set of features used earlier in the path, $f_i \in F$, and $\lambda \in [0,1]$ is the penalty coefficient. RRF avoids including a new feature $j$, except when the value of $Gain(j)$ is greater than $\max \limits_{i}\big( Gain(f_i)\big)$.

To improve RRF, Guided RRF (GRRF) \cite{deng2013gene} assigns a different penalty coefficient $\lambda_{j}$ to each feature instead of assigning the same penalty coefficient to all features. GRRF employs the importance score from a pre-trained RF on the training set to refine the selection of features at a given node. The importance score of feature $j$ in an RF with $T$ trees is the mean of gain for features in the RF. The important scores evaluate the contribution of features for predicting classes. The GRRF uses the normalized importance score to control the degree of regularization of the penalty coefficient as follows:
\begin{equation} \label{14}
\lambda_j = (1-\gamma)\lambda_{0} + \gamma Imp'_{j},
\end{equation}
\noindent where $\lambda_{0} \in (0, 1]$ is the base penalty coefficient and $\gamma \in [0, 1]$ controls the weight of the normalized importance score. The GRRF and RRF are computationally inexpensive methods that are able to select stronger features and avoid redundant features.

\subsection{Dropout}

As detailed in Section \ref{sec:dropout}, dropout is a method that prevents DNNs from overfitting by randomly dropping nodes during the training. Dropout can be added to other machine learning algorithms through two methods: by dropping features or by dropping models in the case of ensemble methods. Dropout has also been employed by dropping input features during training \cite{wang2012baselines} \cite{wang2013fast}. Here we look at techniques that have been investigated for dropping input features, particularly in non-network classifiers. 

Rashmi et al. \cite{vinayak2015dart} applied dropout to Multiple Additive Regression Trees (MART) \cite{friedman2001greedy} \cite{friedman2002stochastic}. MART is a regression tree ensemble which iteratively refines its model by continually adding trees that fit the loss function derivatives from the previous version of the ensemble. Because trees added at later iterations may only impact a small fraction of the training set and thus over-specialize, researchers previously used shrinkage to exclude a random subset of leaf nodes during each tree-adding step. More recently, Rashmi et al. integrated the deep-learning idea of dropout into MART. Using dropout, a subset of the trees are temporarily dropped. A new tree is created based on the loss function for the on-dropped trees. This new tree is combined with the previously-dropped trees into a new ensemble. This method, Dropout Multiple Additive Regression Trees (DART) \cite{vinayak2015dart}, weights the votes for the new and re-integrated trees to have the same effect on the final model output as the original set of trees. Other researchers have experimented with permanently removing a strategic subset of the dropped trees as well \cite{lucchese2017x}.

\subsection{Early Stopping}

The core concept of early stopping is to terminate DNN training once performance on the validation set is not improving. One potential advantage of Deep Forest \cite{zhou2017deep} over DNNs is that DF can determine the depth of a model automatically. In DF, if the model performance does not increase on the validation set after adding a new layer, the learning terminates. Unlike DNNs, DF may avoid the tendency to overfit as more layers are added. Thus, while early stopping does not necessarily enjoy the primary outcome of preventing such overfitting, it can provide additional benefits such as shortening the validation cycle in the search for the optimal tree depth.

\subsection{Data Augmentation}

As discussed in Section \ref{sec:dataaug}, data augmentation is a powerful method for improving DNN generalization. However, little research has investigated the effects of data augmentation methods on non-network classifiers. As demonstrated by Wong et al. \cite{wong2016understanding}, the SVM classifier does not always benefit from data augmentation, in contrast to DNNs. However, Xu \cite{xu2013improvements} ran several data augmentation experiments on synthetic datasets and observed that data augmentation did enhance the performance of random forest classifiers. Offering explanations for the circumstances in which such augmentation is beneficial is a needed area for future research.

\section{Hybrid Models}

Hybrid models can be defined as a combination of two or more classes of models. There are many ways to construct hybrid models, such as DNDF \cite{kontschieder2015deep} which integrates a deep network into a decision forest as explained in Section \ref{sec:optoutofnet}. In this section we discuss other examples of hybrid models.

One motivation for combining aspects of multiple models is to find a balance between classification accuracy and computational cost. Energy consumption by mobile devices and cloud servers is an increasing concern for responsive applications and green computing. Decision forests are computationally inexpensive models because of the conditional property of decision trees. Conversely, while CNNs are less efficient, they can achieve higher accuracy because of their representation-learning capabilities. Ioannou et al. \cite{ioannou2016decision} introduced the Conditional Neural Network (CondNN) to reduce computation in a CNN model by introducing a routing method similar to that found in decision trees. In CondNN, each node in layer $l$ is connected to a subset of nodes from the previous layer, $l-1$. Given a fully trained network, for every two consecutive layers a matrix $\Lambda_{(l-1, l)}$ stores the activation values of these two layers. By rearranging elements of $\Lambda_{(l-1, l)}$ based on highly-active pairs for each class in the diagonal and zeroing out off-diagonal elements, the CondNN develops explicit routes $\Lambda_{(l, l-1)}^{route}$ through nodes in the network. CondNN incurs profoundly lower computation cost compared to other DNNs at test time; whereas, CondNN's accuracy remains similar to larger models. We note that DNN size can be also be reduced by employing Bayesian optimization, as investigated by Blundell et al. \cite{blundell2015weight} and by Fortunato et al. \cite{DBLP:journals/corr/FortunatoBV17}. These earlier efforts provide evidence that Bayesian neural networks are able to decrease network size even more than CondNNs while maintaining a similar level of accuracy.

Another direction for blending a deep network with a non-network classifier is to improve the non-network model by learning a better representation of data via a deep network. Zoran et al. \cite{zoran2017learning} introduce the differentiable boundary tree (DBT) in order to integrate a DNN into the boundary tree \cite{mathy2015boundary} to learn a better representation of data. The newly-learned data representation leads to a simpler boundary tree because the classes are well separated. The boundary tree is an online algorithm in which each node in the tree corresponds to a sample in the training set. The first sample together with its label are established as the tree root. Given a new query sample $z$, the sample traverses through the tree from the root to find the closest node $n$ based on some distance function like the Euclidean distance function. If the label of the nearest node in the tree is different from the query sample, a new node containing the query $z$ is added as a child of the closest node $n$ in the tree; otherwise the query node $z$ is discarded. Therefore, each edge in the tree marks the boundary between two classes and each node tends to be close to these boundaries.

Transitions between nodes in a standard boundary tree are deterministic. DBT combines a SoftMax cost function with a boundary tree, resulting in stochastic transitions. Let $x$ be a training sample and $c$ be the one-hot encoding label of that sample. Given the current node $x_i$ in the tree and a query node $z$, the transition probability from node $x_i$ to node $x_j$, where $x_j \in \{child(x_i), x_i\}$ is the SoftMax of the negative distance between $x_j$ and $z$. This is shown in Equation~\ref{softmax}.
\begin{equation} \label{softmax}
\begin{gathered}
p(x_i \rightarrow x_j|z) = \underset{i,j \in child(i)}{\SoftMax}(-d(x_j,z)) \\ = \frac{\exp(-d(x_j,z))}{\underset{j'\in\{i,j\in child(i)\}}{\sum}\exp(-d(x_j,z))}
\end{gathered}
\end{equation} 
The probability of traversing a particular path in the boundary tree, given a query node $z$, is the product of the probability of each transition along the path from the root to the final node $x_{final^{*}}$ in the tree. The final class log probability of DBT is computed by summing the probabilities of all transitions to the parent of $x_{final^{*}}$ together with the probabilities of the final node and its siblings. The set $sibling(x_i)$ consists of all nodes sharing the same parent with node $x_i$ and the node $x_i$ itself. As discussed earlier, a DNN $f_{\theta}(x)$ transforms the inputs to learn a better representation. The final class log probabilities for the query node $z$ are calculated as follows:
\begin{equation} \label{eq4} 
\begin{gathered}
\log p(c | f_{\theta}(z)) = \smashoperator{\sum\limits_{x_i \rightarrow x_j \in path^{\dagger}|f_{\theta}(z)}} \log p(f_{\theta}(x_i)  \rightarrow f_{\theta}(x_j) | f_{\theta}(z)) \\ + \log \smashoperator{\sum\limits_{x_k \in sibling(x_{final^{*}})}} p(parent(f_{\theta}(x_k)) \rightarrow f_{\theta}(x_k) | f_{\theta}(z)) c(x_k) .
\end{gathered}
\end{equation} 
In Equation~\ref{eq4}, $path^{\dagger}$ denotes $path^{*}$ (the path to the final node $x_{final^{*}}$) without the last transition, and $sibling(x)$ represents node $x$ and all other nodes sharing the same parent with node $x$. The gradient descent algorithm can be applied to Equation \ref{eq4} by plugging in a loss function to learn parameter $\theta$ of the DNN. However, gradient descent cannot be applied easily to DBT because of the node and edge manipulations in the graph. To address this issue, DBT transforms a small subset of training examples via a DNN and builds a boundary tree based on the transformed examples. Next, DBT transforms a query node $z$ via the same DNN and calculates the log probability of a class according to Equation \ref{eq4}. The DNN employs gradient descent to update its parameters by propagating the gradient of log loss probability. DBT discards this boundary tree and iteratively builds a new boundary tree as described until a convergence criteria is met. In the described method, the authors set a specific threshold for the loss value to terminate the training. DBT is able to achieve greater accuracy with a simpler tree than original boundary tree as shown by the authors on the MNIST dataset \cite{lecun1998mnist}. One of the biggest advantages of DBT is its interpretability. However, DBT is computationally an expensive method because a new computation graph needs to be built, which makes batching inefficient. Another limitation is that the algorithm needs to switch between building the tree and updating the tree. Therefore, scaling to large datasets is fairly prohibitive.

Yet another way of building a hybrid model is to learn a new representation of data with a DNN, then hand the resulting feature vectors off to other classifiers to learn a model. Tang \cite{tang2013deep} explored replacing the last layer of DNNs with a linear SVM for classification tasks. The activation values of the penultimate layer are fed as input to an SVM with a $L_2$ regularizer. The weights of the lower layer are learned through momentum gradient descent by differentiating the SVM objective function with respect to activation of the penultimate layer.  The author's experiments on the MNIST \cite{lecun1998mnist} and CIFAR-10 \cite{cifar10} datasets demonstrate that replacing a CNN's SoftMax output layer with SVM yields a lower test error. Tang et al. \cite{tang2013deep} postulate that the performance gain is due to the superior regularization effect of the SVM loss function.

It is worth mentioning that in their experiment on MNIST \cite{lecun1998mnist}, Tang first used PCA to reduce the features and then fed the reduced feature vectors as input to their model. Also, Niu et al. \cite{niu2012novel} replaced the last layer of a CNN with an SVM which similarly resulted in lowering test error of the model compare to a CNN on the MNIST dataset. Similar to these methods, Zareapoor et al. \cite{zareapoor2017kernelized},  Nagi et al. \cite{nagi2012convolutional}, Bellili et al. \cite{bellili2001hybrid}, and Azevedo et al. \cite{azevedo2011mlp} replace the last layer of a DNN with an SVM. In these cases, their results from multiple datasets reveal that employing a SVM as the last layer of a neural network can improve the generalization of the network.

Zhao et al. \cite{zhao2018embedding} replace the last layer of a deep network with a visual hierarchical tree to learn a better solution for image classification problems. A visual hierarchical tree with $L$ levels organizes $N$ objects classes based on their visual similarities in its nodes. Deeper in the tree, groups become more separated wherein each leaf node should contain instances of one class. The class similarity between the class $c_i$ and $c_j$ is defined as follows:
\begin{equation}
    S_{i,j} = S(c_i, c_j) = \exp\Big(-\frac{d(x_i, x_j)}{\sigma} \Big).
\end{equation}
Here, $d(x_i, x_j)$ represents the distance between the deep representation of instances of classes $c_i$ and $c_j$, and $\sigma$ is automatically determined by a self-tuning technique. After calculating matrix $S$, hierarchical clustering is employed to learn a visual hierarchical tree.

In a traditional visual hierarchical tree, some objects might be assigned to incorrect groups. A level-wise mixture model (LMM) \cite{zhao2018embedding} aims to improve this visual hierarchical tree by learning a new representation of data via a DNN then updating the tree during training. For a given tree, matrix $\Psi_{y_i,t_i}$ denotes the probability of objects with label $y$ belonging to group $t$ in the tree. First, LMM updates the DNN parameters and the visual hierarchical tree as is done with a traditional DNN. The only difference is a calculation of two gradients, one based on the parameters of the DNN and other one based on the parameters of the tree. Second, LMM updates the matrix $\Psi_{y_i,t_i}$ for each training sample separately and updates the parameters of the DNN and the tree afterwards. To update the $\Psi$, the posterior probability of the assigning group $t_i$ for the object $x_i$ is calculated based on the number of samples having the same label $y$ as the label of $x_i$ in group $t$. For a given test image, LMM learns a new representation of the image based on the DNN and then obtains a prediction by traversing the tree. One of the advantages of a LMM is that over time, by learning a better representation of data via DNN, the algorithm can update the visual hierarchical tree.

In some cases, two different data views are available. As an example, one view might contain video and the another sound. Canonical correlation analysis (CCA) \cite{hotelling1992relations} and kernel canonical correlation analysis (KCCA) \cite{hardoon2004canonical} offer standard statistical methods for learning view representations that each the most predictable by the other view. Nonlinear representations learned by KCCA can achieve a higher correlation than linear representations learned by CCA. Despite the advantages of KCCA, the kernel function faces some drawbacks. Specifically, the representation is bound to the fixed kernel. Furthermore, because is model is nonparametric, training time as well as the time to compute the new data representation scales poorly with the size of a training set.

Andrews et al. \cite{andrew2013deep} proposed to apply deep networks to learn a nonlinear data representation instead of employing a kernel function. Their resulting deep canonical correlation analysis (DCCA) consists of two separate deep networks for learning a new representation for each view. The new representation learned by the final layer of networks $H_1$ and $H_2$ is fed to CCA. To compute the objective gradient of DCCA, the gradient of the output of the correlation objective with respect to the new representation can be calculated as follows:
\begin{equation}\label{eq:DCCA}
    \frac{\partial_{corr}(H_1,H_2)}{\partial H_1}
\end{equation}
After this computation, backpropagation is applied to find the gradient with respect to all parameters. The details of calculating the gradient in Equation \ref{eq:DCCA} are provided by the authors \cite{andrew2013deep}.

While researchers have also created LSTM methods that employ tree structures \cite{tai2015improved} \cite{alvarez2016tree}, these methods utilize the data structure to improve a network model rather than employing tree-based learning algorithms. Similarly, other researches integrate non-network classifiers into a network structure. Cimino et al. \cite{cimino2016tandem} and Agarap \cite{agarap2018neural} introduce hybrid models. These two methods apply LSTM and GRU, respectively, to learn a network representation. Unlike traditional DNNs, the last layer employs a SVM for classification.

The work surveyed in this section provides evidence that deep neural nets are capable methods for learning high-level features. These features, in turn, can be used to improve the modeling capability for many types of supervised classifiers. 
\begin{table}[h] 
    \centering
    \caption{Summary of classifiers which integrate deep network components into non-network classifiers.}
    \label{tab:my_label}
    \resizebox{\linewidth}{!}{\begin{tabular}{c l l} \toprule[2pt]
         & {\bf Methods} & {\bf Classifiers} \\ \midrule[1pt]
        \multirow{5}{*}{Architecture} & Feedforward & ANT \cite{tanno2018adaptive}, DNDT \cite{yang2018deep}, DBN \cite{hinton2006fast}, Deep PCA \cite{liong2013face}, DF \cite{zhou2017deep}, DPG \cite{damianou2015deep}, R2-SVM \cite{vinyals2012learning}, D-SVM \cite{abdullah2009ensemble}, DTA-LS-SVM \cite{wang2017deep}, SFDT \cite{frosst2017distilling}  \\ \cmidrule{2-3}
        & Autoencoder & DKF \cite{Krishnan2015kalman}, eForest \cite{feng2017autoencoder}, ML-SVM \cite{wiering2014multi} \\ \cmidrule{2-3}
        & Siamese Model & SDF \cite{utkin2017siamese} \\ \cmidrule{2-3}
        & Generative Adversarial Model & GAF \cite{zuo2018generative} \\ \midrule[1pt]
        Optimization & Gradient Decent & DNDF \cite{kontschieder2015deep}, mGBDT \cite{feng2018multi}, ML-SVM \cite{wiering2014multi} \\ \midrule[1pt]
        \multirow{3}{*}{Regularization} & Parameter Norm Penalty & NLP-SVM \cite{fung2004feature}, GRRF \cite{deng2013gene}, RRF \cite{deng2012feature} , SCAD-SVM \cite{zhang2005gene} \\ \cmidrule{2-3}
        & Dropout & DART \cite{vinayak2015dart} \\  \midrule[1pt]
        Hybrid Model & & CondCNN \cite{ioannou2016decision}, DBT \cite{zoran2017learning}, DCCA \cite{andrew2013deep}, DNDF \cite{kontschieder2015deep}, DNN+SVM \cite{tang2013deep} \cite{niu2012novel} \cite{zareapoor2017kernelized} \cite{nagi2012convolutional} \cite{bellili2001hybrid} \cite{azevedo2011mlp}, LMM \cite{zhao2018embedding} \\ \bottomrule[2pt]
    \end{tabular}}
    \label{table:summary}
\end{table}
\bigbreak
In this survey, we aim to provide a thorough review of non-network models that utilize the unique features of deep network models. Table \ref{table:summary} provides a summary of such non-network models, organized based on four aspects of deep networks: model architecture, optimization, regularization, and hybrid model fusing. A known advantage of traditional deep networks compared with non-network models has been the ability to learn a better representation of input features. Inspired by various deep network architectures, deep learning of non-network classifiers has resulted in methods to also learn new feature representations. Another area where non-network classifiers have benefited from recent deep network research is applying backpropagation optimization to improve generalization. This table summarizes published efforts to apply regularization techniques that improve neural network generalization. The last category of models combines deep network classifiers and non-network classifiers to increase overall performance.

\section{Experiments} \label{sec:experiments}

In this paper, we survey a wide variety of models and methods. Our goal is to demonstrate that diverse types of models can benefit from deep learning techniques. To highlight this point, we empirically compare the performance of many techniques described in this survey. This comparison includes deep and shallow networks as well as and non-network learning algorithms. Because of the variety of classifiers that are surveyed, we organize the comparison based on the learned model structure. 

First, we compare the models that are most similar to DNNs. These models should be able to learn a better representation of data when a large dataset is available. We report the test error provided by the authors for MNIST and CIFAR-10 dataset in Table \ref{tab:classifier_perf}. If the performance of a model was not available for any of these datasets, we ran that experiment with the authors' code. Default parameters are employed for parameter values that are not specified in the original papers. In the event that the authors did not provide their code, we did not report any results.
These omissions prevent the report of erroneous performances that result from implementation differences.

The MNIST dataset has been a popular testbed dataset for comparing model choices within the computer vision and deep network communities. MNIST instances contain $28 \times 28$ pixel grayscale images of handwritten digits and their labels. The MNIST labels are drawn from 10 object classes, with a total of 6000 training samples and 1000 testing samples. The CIFAR-10 is also a well-known dataset containing 10 object classes with approximately 5000 examples per class, where each sample is a $32 \times 32$ pixel RGB image. 
\begin{table}[h]
    \centering
    \caption{Classification error rate ($\%$) comparison.}
    \resizebox{\linewidth}{!}{\begin{tabular}{l c c c c c c c c c} \toprule[2pt]
        Dataset/{\bf Models} & {\bf RF} & {\bf ANT$\dagger$} \cite{tanno2018adaptive} & {\bf DF} \cite{zhou2017deep} & {\bf R2SVM} \cite{vinyals2012learning} & {\bf SFDT} \cite{frosst2017distilling} & {\bf DNDF} \cite{kontschieder2015deep} & {\bf CondCNN} \cite{ioannou2016decision} & {\bf DBT} \cite{zoran2017learning} & {\bf DNN+SVM} \cite{tang2013deep}\\ \midrule
        MNIST & 3.21 & 0.29 & 0.74 & 4.03$^*$ & 5.55 & 0.7 & - & 1.85 & 0.87 \\
        CIFAR-10 & 50.17 & 7.71 & 31.0 & 20.3 & - & - & 10.12 & 13.06 & 11.9\\
        \bottomrule[2pt]
    \end{tabular}}
    \begin{tablenotes}
      \small
      \item * Based on the authors' code.
      \item $\dagger$ The reported result reflects an ANT ensemble.
    \end{tablenotes}
    \label{tab:classifier_perf}
\end{table}{}

In the next set of experiments, we compare models designed to handle small or structured datasets, as shown in Table  \ref{tab:classifier_small_perf}. The authors of these methods tested a wide range of datasets to evaluate their their approach. In this survey, we conducted a series of experiments on the UCI human activity recognition (HAR) dataset \cite{anguita2013public}. Here, human activity recognition data were collected from 30 participants performing six scripted activities (walking, walking upstairs, walking downstairs, sitting, standing, and laying) while wearing smartphones. The dataset contains 561 features extracted from sensors including an accelerometer and a gyroscope. The training set contains 7352 samples from 70\% of the volunteers and the testing set contains 2947 samples from the remaining 30\% of volunteers.
\begin{table}[h]
    \centering
    \caption{Classification error rate ($\%$) comparison.}
    \resizebox{\columnwidth}{!}{\begin{tabular}{l c c c c c c c} \toprule[2pt]
        Dataset/{\bf Models} & {\bf RF} & {\bf SVM} & {\bf MLP} & {\bf DART} \cite{vinayak2015dart} & {\bf RRF} \cite{deng2012feature}  & {\bf GRRF} \cite{deng2013gene} & {\bf mGBDT} \cite{feng2018multi} \\ \midrule
        HAR & 6.96 & 4.69 & 4.69 & 6.55 & 3.77 & 3.74 & 7.68 \\
        \bottomrule[2pt]
    \end{tabular}}
    \label{tab:classifier_small_perf}
\end{table}{}

From the table, we observe that models representing multiple layers of machine learning models such as DF, R2SVM, and mGBDT did not perform well on the MNIST and CIFAR-10 datasets. Compared to DNNs, these models are computationally more expensive, require an excessive amount of resources, and do not offer advantages of interpretability.

Another category of models utilizes a neural network to learn a better representation of data. These models such as DNDF, and DNN+SVM applied a more traditional machine learning model on the newly-learned representation. This technique could be beneficial when the neural network has been trained on a large dataset. For example, DNDF utilized GoogLeNet for extracting features, and subsequently achieved a $6.38\%$ error rate on the ImageNet testset. In contrast, the GoogLeNet error rate is $10.02\%$. Another class of models enhances the tree model by integrating artificial neurons such as ANT, SFDT, and DBT. These models cleverly combine neural networks with decision trees that improve interpretation while offering representation-learning benefits.

Another hybrid strategy focused on decreasing the computational complexity of the DNNs. CondCNN is such a neural network that employs a routing mechanism similar to a decision tree to achieve this goal. Another successful line of research is to add regularizers frequently used by the neural network to other classifiers similar to DART, RRF, and GRRF.

The results from our experiments reveal that both network classifiers and non-network classifiers benefit from deep learning. The methods surveyed in this paper and evaluated in these experiments demonstrate that non-network machine learning models do improve performance by
incorporating DNN components into their algorithms. Whereas models without feature learning such as RF usually do not perform well on unstructured data such as images, we observe that adding deep learning to these models drastically improve their performance, as shown in Table \ref{tab:classifier_perf}. Additionally, non-deep models may achieve improved performance on structured data by adding regularizers, as shown in Table \ref{tab:classifier_small_perf}. The methods surveyed in this paper demonstrate that deep learning components can be added to any type of machine learning model, and are not specific to DNNs. The incorporation of deep learning strategies is a promising direction for all types of classifiers, both network and non-network methods.

\section{Conclusions and Directions for Ongoing Research}

DNNs have emerged as a powerful force in the machine learning field for the past few years. This survey paper reviews the latest attempts to incorporate methods that are traditionally found in DNNs into other learning algorithms. DNNs work well when there is a large body of training data and available computational power. DNNs have consistently yielded strong results for a variety of datasets and competitions, such as winning the Large Scale Visual Recognition Challenge \cite{ilsvrc2017} and achieving strong results for energy demand prediction \cite{paterakis2017deep}, identifying gender of a text author \cite{sboev2018deep}, stroke prediction \cite{hung2017comparing}, network intrusion detection \cite{yin2017deep}, speech emotion recognition \cite{fayek2017evaluating}, and taxi destination prediction \cite{de2015artificial}. Since there are many applications which lack large amounts of training data or for which the interpretability of a learned model is important,  there is a need to integrate the benefits of DNNs with other classifier algorithms. Other classifiers have demonstrated improved performance on some types of data, therefore the field can benefit from examining ways of combining deep learning elements between network and non-network methods.

Although some work to date provides evidence that DNN techniques can be used effectively by other classifiers, there are still many challenges that researchers need to address, both to improve DNNs and to extend deep learning to other types of classifiers. Based on our survey of existing work, some related areas where supervised learners can benefit from unique DNN methods are outlined below. 

The most characteristic feature of DNNs is a deep architecture and its ability to learn a new representation of data. A variety of stacked generalization methods have been developed to allow other machine learning methods to utilize deep architectures as well. These methods incorporate multiple classification steps in which the input of the next layer represents the concatenation of the output of the previous layer and the original feature vector as discussed in Section \ref{sec:outfeedforward}. Future work can explore the many other possibilities that exist for refining the input features to each layer to better separate instances of each class at each layer. 

Previous studies provide evidence that DNNs are effective data generators \cite{radford2015unsupervised} \cite{hoffman2017cycada}, while in some cases non-network classifiers may actually be the better discriminators. Future research can consider using a DNN as a generator and an alternative classifier as a discriminator in generative adversarial models. Incorporating this type of model diversity could improve the robustness of the models.

Gradient descent can be applied to any differentiable algorithm. We observed that Kontschieder et al. \cite{kontschieder2015deep}, Frosst et al. \cite{frosst2017distilling}, Tanno et al. \cite{tanno2018adaptive}, and Zoran et al. \cite{zoran2017learning} all applied gradient descent to two different tree-based algorithms by making them differentiable. In the future, additional classifiers can be altered to be differentiable. Applying gradient descent to other algorithms could be an effective way to adjust the probability distribution of parameters.

Another area which is vital to investigate is the application of network-customized regularization methods unique to non-network classifiers. As discussed in Section \ref{sec:regoutdnn}, the non-network classifiers can benefit from the regularization methods that are unique to DNNs. However, there exist many different ways that these regularization methods can be adapted by non-network classifiers to improve model generalization.

An important area of research is interpretable models. There exist applications such as credit score, insurance risk, health status because of their sensitivity, models need to be interpretable. Further research needs to exploit the use of DNNs in interpretable models such as DNDT \cite{yang2018deep}.

As we discussed in this survey, an emerging area of research is to combine the complementary benefits of statistical models with neural networks. Statistical models offer mathematical formalisms as well as possible explanatory power. This combination may provide a more effective model than either approach used in isolation.

There are cases in which the amount of ground truth-labeled data is limited, but a large body of labeled data from the same or similar distribution is available. One possible area of ongoing exploration is to couple the use of DNNs for learning from unlabeled data with the use of other classifier strategies for learning from labeled data. The simple model learned from labeled data can be exploited to further tune and improve learned representation patterns in the DNN.

We observe that currently, there is a general interest among the machine learning community to transfer new deep network developments to other classifiers. While a substantial effort has been made to incorporate deep learning ideas into the general machine learning field, continuing this work may spark the creation of new learning paradigms. However, the benefit between network-based learners and non-network learners can be bi-directional. Because a tremendous variety of classifiers has shown superior performance for a wide range of applications, future research can focus not only on how DNN techniques can improve non-network classifiers but on how DNNs can incorporate and benefit from non-network learning ideas as well.
\begin{table}[h]
    \centering
    \caption{The list of abbreviations and their descriptions utilized in this survey.}
    \label{tab:abb_dis}
    \begin{tabular}{l l} \toprule[2pt]
        {\bf Abbreviation} & {\bf Description} \\ \midrule[1pt]
        AE & Autoencoder \\
        ANT & Adaptive Neural Tree \\
        CNN & Convolutional  Neural  Network \\ 
        CondNN & Conditional Neural Network \\
        DART & Dropout Multiple Additive Regression Trees \\
        DBT & Differentiable Boundary Tree \\
        DBN & Deep Belief Network \\
        DCCA & Deep Canonical Correlation Analysis \\ 
        Deep PCA &  Deep principal components analysis \\
        DF & Deep Forest \\
        DGP & Deep Gaussian Processes \\
        DKF & Deep Kalman Filters \\
        DNDT & Deep Network Decision Tree  \\
        DNDF & Deep Network Decision Forest  \\
        DNN & Deep Neural Network \\
        DSVM & Deep SVM \\
        DTA-LS-SVM & Deep Transfer Additive Kernel Least Square SVM \\
        eForest & Encoder Forest \\
        FC & Fully Connected \\ 
        GAF & Generative Adversarial Forest \\
        GAN & Generative Adversarial Network \\
        GRRF & Guided Regularized Random Forest \\
        LMM & Level-wise Mixture Model \\
        mGBDT & Multilayer Gradient Decision Tree \\
        ML-SVM & Multilayer SVM \\
        MLP & Multilayer perceptron  \\
        NLP-SVM &  Newton Linear Programming SVM \\
        R2-SVM & Random Recursive SVM \\
        RBM & Restricted Boltzmann Machine \\
        RNN & Recurrent Neural Network \\
        RRF & Regularized Random Forest \\
        SCAD-SVM & Smoothly Clipped Absolute Deviation SVM \\
        SDF & Siamese Deep Forest \\
        SNN & Siamese Neural Network \\
        FSDT & Frosst Soft Decision Tree \\
        VAE & Variational Autoencoder \\
        \bottomrule[2pt]
    \end{tabular}
\end{table}

\section*{Acknowledgment}

The authors would like to thank Tharindu Adikari, Chris Choy, Ji Feng, Yani Ioannou, Stanislaw Jastrzebski and Marco A. Wiering for their valuable assistance in providing code and additional implementation details of the algorithms that were evaluated in this paper. We would also like to thank Samaneh Aminikhanghahi and Tinghui Wang for their feedback and guidance on the methods described in this survey. This material is based upon work supported by the National Science Foundation under Grant No. 1543656.

\bibliographystyle{unsrt}
\bibliography{references}

\end{document}